\documentclass{article}
\usepackage{ijcai26}
\usepackage{times}
\usepackage[utf8]{inputenc}
\usepackage{graphicx} \graphicspath{{figures/}}
\usepackage{amsmath,amssymb,mathabx,mathtools,amsthm,nicefrac}
\usepackage[hidelinks]{hyperref}

\usepackage[skip=3pt,font=small]{subcaption}
\usepackage[skip=3pt,font=small]{caption}
\usepackage{acronym}
\usepackage{wrapfig}
\usepackage[dvipsnames,svgnames,x11names,table]{xcolor}
\usepackage{enumerate}
\usepackage{xspace}
\usepackage[misc]{ifsym}
\usepackage{titletoc}
\usepackage{enumitem}
\usepackage{verbatim,fancyvrb,listings}
\usepackage{booktabs,tabularx,colortbl,multirow,multicol,array,makecell}
\usepackage{siunitx}
\usepackage[most]{tcolorbox}
\usepackage{textcomp} 
\usepackage[normalem]{ulem}
\usepackage[switch]{lineno}
\usepackage[capitalise,noabbrev,nameinlink]{cleveref}

\definecolor{grey}{RGB}{190,190,190}

\newtcolorbox{promptbox}[1]{
    enhanced,
    breakable,
    colback=grey!15!white,         
    colframe=black!75!white,       
    coltitle=white,                
    fonttitle=\sffamily\bfseries,
    fontupper=\small\ttfamily,     
    arc=0.5mm,                     
    boxrule=0.5pt,                 
    borderline west={1.5pt}{0pt}{black!75!white},
    borderline east={1.5pt}{0pt}{black!75!white},
    borderline south={1.5pt}{0pt}{black!75!white},
    titlerule=0pt,                 
    left=15pt, right=15pt,         
    top=10pt, bottom=10pt,
    before skip=15pt,
    after skip=15pt,
    title=#1
}

\newtcolorbox{inlinepromptbox}[1]{
    enhanced,
    breakable,
    colback=grey!15!white,         
    colframe=black!75!white,       
    coltitle=white,                
    fonttitle=\sffamily\bfseries,
    fontupper=\small\ttfamily,     
    arc=0.5mm,                     
    boxrule=0.5pt,                 
    borderline west={1.5pt}{0pt}{black!75!white},
    borderline east={1.5pt}{0pt}{black!75!white},
    borderline south={1.5pt}{0pt}{black!75!white},
    titlerule=0pt,                 
    left=10pt, right=10pt,         
    top=5pt, bottom=5pt,
    before skip=15pt,
    after skip=15pt,
    title=#1
}

\makeatletter
\DeclareRobustCommand\onedot{\futurelet\@let@token\@onedot}
\def\@onedot{\ifx\@let@token.\else.\null\fi\xspace}
\def\eg{\emph{e.g}\onedot}

\def\etc{\emph{etc}\onedot} 
\def\vs{\emph{vs}\onedot}

\makeatother

\acrodef{llm}[LLM]{Large Language Model}
\acrodef{ai}[AI]{Artificial Intelligence}
\acrodef{ood}[OOD]{Out-of-Distribution}

\title{BotzoneBench: Scalable LLM Evaluation via Graded AI Anchors}

\author{
Lingfeng Li$^1$\and
Yunlong Lu$^1$\and
Yuefei Zhang$^2$\and
Jingyu Yao$^3$\and
Yixin Zhu$^4$* \and
KeYuan Cheng$^1$ \and
Yongyi Wang$^1$ \and
Qirui Zheng$^1$ \and
Xionghui Yang$^1$ \and
Wenxin Li$^1$*\\
\affiliations
$^1$School of Computer Science, Peking University \\
$^2$School of Software Engineering, South China University of Technology \\
$^3$Department of Computer Science, Yale University \\
$^4$School of Psychological and Cognitive Sciences, Peking University\\
\emails
*Correspondence to: Yixin Zhu (yixin.zhu@pku.edu.cn),
Wenxin Li (lwx@pku.edu.cn)
}

\begin{document}
\maketitle

\begin{abstract}
	\acp{llm} are increasingly deployed in interactive environments requiring strategic decision-making, yet systematic evaluation of these capabilities remains challenging.
	Existing benchmarks for \acp{llm} primarily assess static reasoning through isolated tasks and fail to capture dynamic strategic abilities. Recent game-based evaluations employ \acs{llm}-vs-\acs{llm} tournaments that produce relative rankings dependent on transient model pools, incurring quadratic computational costs and lacking stable performance anchors for longitudinal tracking.
	The central challenge is establishing a scalable evaluation framework that measures \acs{llm} strategic reasoning against consistent, interpretable standards rather than volatile peer models.
	Here we show that anchoring \ac{llm} evaluation to fixed hierarchies of skill-calibrated game \ac{ai} enables linear-time absolute skill measurement with stable cross-temporal interpretability.
	Built on the Botzone platform's established competitive infrastructure, our BotzoneBench evaluates \acp{llm} across eight diverse games spanning deterministic perfect-information board games to stochastic imperfect-information card games. Through systematic assessment of 177,047 state-action pairs from five flagship models, we reveal significant performance disparities and identify distinct strategic behaviors, with top-performing models achieving proficiency comparable to mid-to-high-tier specialized game \ac{ai} in multiple domains.
	This anchored evaluation paradigm generalizes beyond games to any domain with well-defined skill hierarchies, establishing a scalable and reusable framework for assessing interactive \ac{ai} capabilities.
\end{abstract}

\section{Introduction}

As \acp{llm} continue to advance, researchers have developed increasingly diverse benchmarks to evaluate their emergent capabilities~\cite{moradi2025critical}. Static benchmarks such as MATH~\cite{hendrycks2021measuring} for mathematical reasoning, HumanEval~\cite{chen2021evaluatinglargelanguagemodels} for code generation, and MMLU~\cite{hendrycks2020measuring} for general knowledge have proven valuable for assessing isolated, single-turn reasoning. However, they inherently fail to capture \acp{llm} performance in dynamic, interactive decision-making settings. Games, with their well-defined rules, clear objectives, and demand for strategic planning, offer a natural testbed for probing multi-step reasoning, long-horizon planning, and adaptive behavior under uncertainty~\cite{Bellemare_2013,silver2016mastering,vinyals2019grandmaster}.

\begin{figure}[t!]
	\centering
	\includegraphics[width=\linewidth]{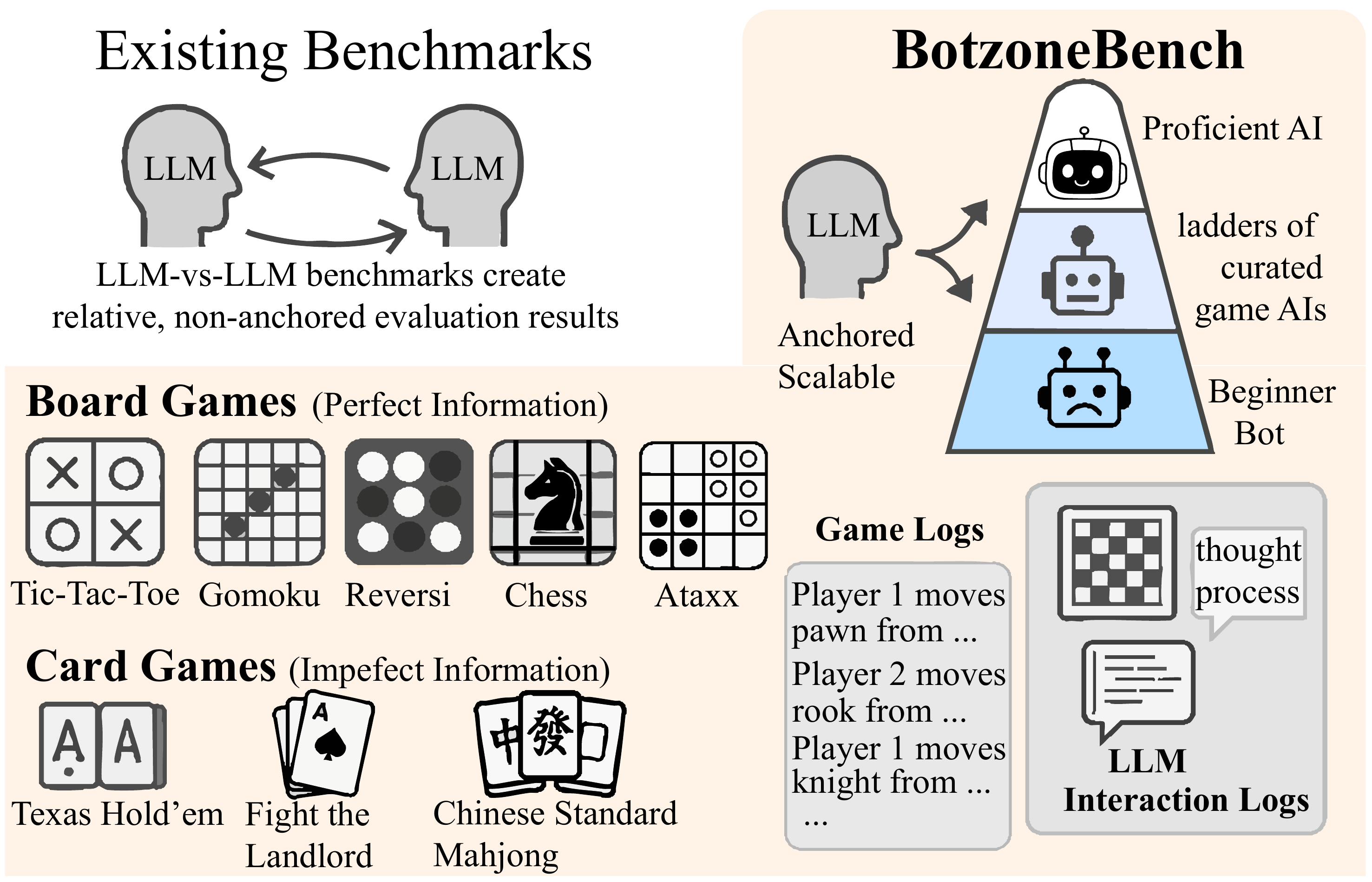}
	\caption{\textbf{Anchored evaluation in BotzoneBench compared to existing game benchmarks.} Existing benchmarks employ \acs{llm}-vs-\acs{llm} tournaments requiring $O(N^2)$ pairwise evaluations with relative rankings. BotzoneBench anchors evaluation to fixed skill-calibrated bot hierarchies, enabling $O(N)$ scalable assessment with absolute ratings across eight games spanning deterministic perfect-information (Tic-Tac-Toe, Gomoku, Reversi, Chess, Ataxx) and stochastic imperfect-information settings (Texas Hold'em, Fight the Landlord, Mahjong).}
	\label{fig:intro}
\end{figure}

Recently, game-based benchmarks have emerged to assess the interactive abilities of \acp{llm}~\cite{costarelli2024gamebench,huang2025competing,guo2025r,gui2025logicgame}. These approaches typically employ \acs{llm}-vs-\acs{llm} tournaments to evaluate strategic reasoning, spatial understanding, and opponent modeling. While they provide initial insights, they suffer from three fundamental limitations: (i) \emph{relative rankings} dependent on transient model pools lack absolute interpretability; (ii) \emph{quadratic computational cost} ($O(N^2)$ pairwise matches) hinders scalability; and (iii) \emph{unstable baselines} make longitudinal tracking of model progress across generations infeasible.

To address these challenges, we introduce \textbf{BotzoneBench}, a scalable evaluation framework built on the mature Botzone platform\footnote{Botzone (\url{https://www.botzone.org.cn}) hosts over 23,000 registered users, 213,000 \ac{ai} bots, and 7 million matches played.}. As illustrated in \cref{fig:intro}, BotzoneBench employs an \textit{anchored evaluation} methodology: \acp{llm} are assessed against a fixed hierarchy of skill-calibrated, non-\acs{llm} game \ac{ai} bots rather than volatile peer models. This paradigm shift yields five key contributions:
\begin{itemize}[leftmargin=*,noitemsep,nolistsep]
	\item \textbf{Diverse game portfolio}: Eight games spanning deterministic perfect-information board games (\eg, Chess, Reversi) to stochastic imperfect-information card games (\eg, Mahjong, Texas Hold'em), systematically varying determinism, information availability, and strategic complexity for comprehensive profiling of \ac{llm} reasoning capabilities.
    \item \textbf{Anchored evaluation paradigm}: Graded classic \ac{ai} bots provide stable performance anchors, reducing evaluation complexity from $O(N^2)$ to $O(N)$, enabling efficient model onboarding, and supporting longitudinal cross-family comparisons without benchmark recalibration.
    \item \textbf{Balanced prompting strategy}: A framework that provides necessary game information (state representation, legal actions) to ensure rule compliance while deliberately avoiding prescriptive reasoning templates, prioritizing assessment of genuine strategic decision-making over prompt engineering artifacts.
    \item \textbf{Large-scale annotated dataset}: 177,047 state-action pairs from 6,403 games, each annotated with model reasoning traces, selected actions, and environmental feedback, enabling detailed analysis of \ac{llm} decision-making, strategic adaptation, and error patterns.    
    \item \textbf{Systematic empirical analysis}: Evaluation of five flagship \acp{llm} reveals that top-performing models (notably Gemini) achieve proficiency comparable to mid-to-high-tier specialized game \ac{ai}, while tests across the Qwen3 family (8B---235B parameters) confirm that strategic performance scales predictably with model size.
\end{itemize}

Beyond games, our anchored evaluation paradigm generalizes to any domain with well-defined skill hierarchies---robotic control, dialogue systems, or structured decision-making tasks. While we retain optional \acs{llm}-vs-\acs{llm} modes for supplementary analysis, our primary contribution is a scalable, reusable framework enabling rigorous longitudinal studies of strategic reasoning in \acp{llm}\footnote{All code and data are publicly available at \href{https://github.com/AMysteriousBeing/BotzoneBench}{https://github.com/AMysteriousBeing/BotzoneBench}.}.

\section{Related Work}

Our work establishes skill-graded performance anchors for \acp{llm} in game-based evaluation. We review three lines of research: general-purpose \ac{llm} benchmarks that contextualize the static evaluation paradigm, game-based \ac{llm} benchmarks that introduce interactive assessment yet rely on relative rankings, and classic game \ac{ai} platforms that offer precise skill calibration and mature competitive infrastructure.

\paragraph{General-purpose \texorpdfstring{\acs{llm}}{} benchmarks}

\acp{llm} have spurred numerous benchmarks measuring performance across cognitive and technical tasks in isolated, static settings. Mathematical reasoning is assessed through GSM8K~\cite{cobbe2021training} and MATH~\cite{hendrycks2021measuring}, code generation via HumanEval~\cite{chen2021evaluatinglargelanguagemodels} and MBPP~\cite{austin2021program}, and broad knowledge through MMLU~\cite{hendrycks2020measuring} and Big-Bench Hard~\cite{suzgun2023challenging}. Agentic capabilities are probed by AgentBench~\cite{liu2023agentbench} and WebArena~\cite{zhou2023webarena}, assessing tool use and multi-step planning.
However, these static paradigms suffer from two critical limitations. First, they fail to probe long-horizon strategic planning or real-time adaptive interaction. Second, they are susceptible to data leakage and test contamination~\cite{zhou2023dontmakellmevaluation}, where models exploit memorized patterns rather than demonstrate genuine reasoning. These inherent constraints motivate specialized game-based benchmarks for evaluating strategic decision-making.

\paragraph{Game-based benchmarks for \texorpdfstring{\acsp{llm}}{}}

To evaluate strategic and interactive capabilities, researchers have adopted classic games as dynamic testbeds. GamaBench~\cite{huang2025competing} employs game-theoretic models with self-play, while GameBench~\cite{costarelli2024gamebench} expands to board and communication games with multi-agent interaction over 20-50 rounds. GameBot~\cite{lin2025gamebot} and LLMArena~\cite{chen2024llmarena} cover broader game types (board, card, arcade) and emphasize prompting techniques through \acs{llm}-vs-\acs{llm} tournaments. LVLM-Bench~\cite{wang2025large} adopts visual elements with Minimax opponents, relying on completion-based metrics rather than competitive proficiency.

While these benchmarks pioneer interactive evaluation and reveal insights into spatial reasoning and opponent modeling, they share a \textbf{fundamental limitation}: assessments remain intrinsically relative, dependent on transient model pools, and lack stable absolute anchors for skill calibration. Furthermore, existing approaches exhibit opposing methodological extremes. GameBot heavily structures reasoning via manual Chain-of-Thought prompts, potentially constraining natural strategic thinking, while LVLM-Bench provides minimal guidance and evaluates move counts rather than play quality. 

BotzoneBench addresses both limitations by (i) anchoring evaluation to skill-calibrated classic \ac{ai} bots, yielding stable absolute performance metrics, and (ii) balancing information provision (game state, legal actions) with strategic autonomy, avoiding prescriptive reasoning templates while ensuring rule compliance. Performance is measured through win rates against graded \ac{ai} baselines, enabling robust and interpretable assessment of strategic capability.

\begin{table*}[t!]
	\centering
	\setlength{\tabcolsep}{3pt}
	\small
	\resizebox{\linewidth}{!}{%
    	\begin{tabular}{lcclp{8cm}}
    		\toprule
    		\textbf{Perfect information} & \textbf{Stochasticity} & \multicolumn{2}{c}{\textbf{State space size}} & \textbf{Primary challenge for \acs{llm}}\\
    		\midrule
    		Tic-Tac-Toe & Deterministic & \multicolumn{2}{c}{$\sim10^4$} & Basic rule-following, exhaustive search.\\
    		Gomoku & Deterministic & \multicolumn{2}{c}{$\sim10^{107}$} & Tactical pattern recognition, medium-term planning.\\
    		Ataxx & Deterministic & \multicolumn{2}{c}{$\sim10^{23}$} & Dynamic board, complex mobility and capture tactics.\\
    		Reversi & Deterministic & \multicolumn{2}{c}{$\sim10^{30}$} & Positional evaluation, long-term consequence of moves.\\
    		Chess & Deterministic & \multicolumn{2}{c}{$\sim10^{43}$} & Deep strategic planning, complex piece coordination.\\
    		\midrule
    		\textbf{Imperfect information} & \textbf{Stochasticity} & \textbf{Information set} & \textbf{Information size} & \textbf{Primary challenge for \acs{llm}}\\
    		\midrule
    		Texas Hold'em & Stochastic & $\sim10^{162}$ & $\sim10^{3}$ & Probabilistic reasoning, opponent modeling, bluffing.\\
    		Fight the Landlord & Stochastic & $\sim10^{83}$ & $\sim10^{23}$ & Partial observability, complex coalition dynamics.\\
    		MCR Mahjong & Stochastic & $\sim10^{121}$ & $\sim10^{48}$ & Complex scoring, multi-agent tactics, tile efficiency.\\
    		\bottomrule
    	\end{tabular}%
    }%
    \caption{\textbf{Taxonomy of selected evaluation games.} The table categorizes eight games by information availability (perfect \vs imperfect), stochasticity, and computational complexity. State space size measures perfect-information games, while information set count and average information set size characterize imperfect-information games. The rightmost column highlights key challenges each game poses for \acp{llm}.}
	\label{tab:game_chart}
\end{table*}

\paragraph{Classic game \texorpdfstring{\acs{ai}}{} platforms}

Game \ac{ai} research has been supported by two complementary platform paradigms. \textbf{Local simulation frameworks} such as the Arcade Learning Environment~\cite{Bellemare_2013}, OpenAI Gym (now Gymnasium~\cite{towers2025gymnasiumstandardinterfacereinforcement}), OpenSpiel~\cite{lanctot2019openspiel}, and MineDojo~\cite{fan2022minedojobuildingopenendedembodied} provide developer control and reproducibility, facilitating rapid prototyping, hyperparameter tuning, and deep integration with learning frameworks. However, they generally lack an integrated ecosystem of publicly ranked agents, limiting their ability to provide stable, skill-anchored benchmarks for evaluation.

Conversely, \textbf{online competitive platforms} such as Botzone~\cite{zhou2018botzone} offer diverse card and board game environments with public, stable skill hierarchies. Botzone has been widely adopted in university curricula and serves as the official platform for academic game \ac{ai} competitions at IJCAI~\cite{lu2024mahjong}. However, such platforms typically impose architectural constraints that hinder modern research workflows: iterative debugging requires manual upload cycles, systematic comparative evaluation lacks native support, and the integration of \acs{llm}-based agents is minimal.


BotzoneBench bridges this gap by leveraging Botzone's stable bot hierarchies through a local development environment that combines the structured evaluation of an online platform with the flexibility of local workflows. This architecture enables efficient \ac{llm} benchmarking while maintaining compatibility with Botzone's established skill calibration system.

\section{Selected Games and Rationale}

To assess the strategic reasoning and adaptive capabilities of \acp{llm}, we selected a suite of games from the Botzone platform guided by three principles:
\begin{itemize}[leftmargin=*,noitemsep,nolistsep]
	\item \textbf{Diversity in game-theoretic properties}: state transition stochasticity (deterministic/stochastic), information availability (perfect/imperfect information), and agent count (2-4 players);
	\item \textbf{Gradient in computational complexity}: measured by state space size for perfect-information games, and information set count plus average information set size for imperfect-information games;
	\item \textbf{Availability of graded classic \ac{ai} baselines}: essential for establishing reliable performance anchors.
\end{itemize}

The chosen games, summarized in \cref{tab:game_chart} and detailed below, collectively form a rigorous testbed for probing \ac{llm} capabilities in interactive decision-making. Complexity measures in \cref{tab:game_chart} are drawn from prior studies~\cite{allis1994searching,lu2023official}.

\subsection{Deterministic, Perfect-Information Games}

This category tests pure logical and strategic reasoning in fully observable environments:
\begin{itemize}[leftmargin=*,noitemsep,nolistsep]
	\item \textbf{Tic-Tac-Toe}: A solved game serving as a sanity check for basic rule understanding and turn-taking.
	\item \textbf{Gomoku}: A tactical pattern-matching game focused on forming five-in-a-row. Despite its vast state space, Gomoku without opening restrictions is solved with a known winning strategy~\cite{allis1994searching}, representing moderate strategic complexity. It tests goal-directed planning and defensive blocking.
	\item \textbf{Ataxx}: A dynamic connection/capture game where pieces move or clone. It challenges board mobility management, territorial aggression, and complex local battles.
	\item \textbf{Reversi}: A positional strategy game centered on corner control, testing long-term positional judgment and move consequence evaluation.
	\item \textbf{Chess}: The classic benchmark for strategic depth, testing planning, piece coordination, and balancing tactical combinations with strategic goals.
\end{itemize}

\subsection{Stochastic, Imperfect-Information Games}

This category introduces hidden information and randomness, probing probabilistic reasoning and adaptability:
\begin{itemize}[leftmargin=*,noitemsep,nolistsep]
	\item \textbf{Texas Hold'em (2-Player)}: A canonical imperfect-information game testing probabilistic calculation (pot odds), opponent modeling (betting patterns), and strategic deception (bluffing). Each match consists of 20 hands, with victory determined by final chip count.
	\item \textbf{Fight the Landlord}: A three-player coalitional card game where two \textit{Peasants} cooperate against a single \textit{Landlord}. It introduces multi-agent dynamics, partial observability, and cooperative-competitive strategies, demanding flexible role adaptation and coordinated play.
	\item \textbf{Official International Mahjong}: A multi-player tile-based game characterized by imperfect information, complex scoring, and dynamic player interactions. This version follows Mahjong Competition Rules (MCR)~\cite{li2025mxplainer}, featuring 81 scoring patterns and requiring at least 8 points to win. It serves as a high-ceiling test for multifaceted probabilistic reasoning, memory, and adaptive tactical play under uncertainty.
\end{itemize}

\paragraph{Rationale}

This diverse portfolio ensures our benchmark does not overfit to a single cognitive challenge. The progression from simple perfect-information to complex imperfect-information games enables us to trace how \ac{llm} reasoning capabilities scale with problem complexity in interactive environments. Crucially, each game provides a ladder of pre-calibrated classic \ac{ai} bots, offering stable anchors for absolute (not merely relative) performance assessment.

We also developed \acs{llm}-based bots for \textbf{Go}, although they were excluded from the final benchmark due to the extreme complexity that often led to reasoning breakdowns, timeouts, or invalid decisions in late-game stages.Nevertheless, the \textbf{Go} environment and corresponding \ac{llm} agent have been publicly released to support future research.

\section{Experimental Setup}\label{sec:method}

This section details the experimental protocols for benchmarking \acs{llm}-based game-playing agents, including match design, construction and prompting of \ac{llm} bots, and selection criteria for classic \ac{ai} baselines.

\subsection{Evaluation Protocol: Matches and Seeding}

To ensure robust and statistically meaningful comparisons, we employ a duplicate match design~\cite{quint2007new} that controls for turn-order bias and stochastic variance. Under this design, each evaluated \ac{llm} plays a duplicate set of games against every graded baseline bot, with the \ac{llm} occupying each player position (\eg, first and second mover) an equal number of times.

For perfect-information games, each \acs{llm}-baseline matchup consists of 16 duplicate games per side, totaling 32 games. For imperfect-information games, where randomness significantly affects outcomes, we increase duplicates to account for higher variance. Each matchup extends to 32 duplicate games per side, totaling 64 games.

To ensure direct comparability and full reproducibility across all evaluations, the sequence of random events (\eg, card shuffles) is deterministically controlled by a fixed set of 32 random seeds, applied identically to every bot matchup. This seeded approach eliminates luck as a confounding variable, allowing fair assessment of strategic capability.

To distinguish between transient failures (\eg, API timeouts or network connectivity issues) and genuine strategic inability, we automatically rematch any game that terminates due to such errors from the \acs{llm}-based bot. This ensures that temporary technical interruptions do not conflate with actual deficiencies in gameplay reasoning.

\begin{figure}[t!]
	\centering
	\includegraphics[width=\linewidth]{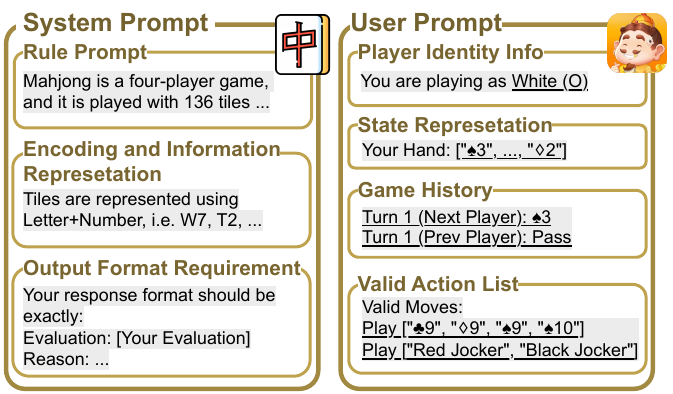}
	\caption{\textbf{Prompt template for \acs{llm}-based bots.} \underline{Underscored} text is dynamically generated for each state. The system prompt provides static game rules and output format requirements, while the user prompt delivers dynamic turn-specific information: player identity, current game state, action history, and legal moves. This design supplies all essential context while leaving strategic reasoning entirely to \acsp{llm}.}
	\label{fig:llm_bot}
\end{figure}

\subsection{Architecture of \acs{llm}-Based Bot}

Our evaluation integrates large language models as interactive game-playing agents. The architecture operates in a sequence that mirrors classic neural network-based agents (state observation, feature extraction, and action masking) while fundamentally differing in state representation: observations are formatted as comprehensive textual prompts rather than numerical tensors, maximizing contextual information for \ac{llm} reasoning.

At each turn, the bot receives a structured prompt containing the game state, legal actions, and, where applicable, action history. This information is processed by the \ac{llm} to produce a reasoned decision. The prompt structure is detailed in \cref{sec:prompting_strat}.

To ensure robust evaluation focused on decision-making, the architecture includes fault-tolerant handling of invalid outputs and timeouts. Each decision is allocated up to 300 seconds; a timeout results in an error, preventing unresponsiveness from being masked by automated recovery.

The \ac{llm}-based bot functions as a reasoning-driven agent that maps textual state representations to actions, enabling consistent and interpretable assessment against skill-calibrated baselines.

\subsection{Prompting Strategy for \texorpdfstring{\acs{llm}}{}-Based Bot}\label{sec:prompting_strat}

This section details the structured prompting strategy employed to transform a \ac{llm} into a functional game-playing agent. The design follows a principled decomposition of prompts into two complementary components: a \textbf{system prompt} that provides static, game-agnostic instructions and constraints, and a \textbf{user prompt} that conveys dynamic, turn-specific information required for decision-making. This separation ensures clarity, reproducibility, and consistent reasoning across all game environments.

As illustrated in \cref{fig:llm_bot}, the \textbf{system prompt} comprises three core elements. The \textbf{rule prompt} provides a precise, textual description of the game's rules, objectives, and key mechanics. Where necessary, concise examples clarify complex rules or common scenarios, ensuring \acp{llm} operate with a correct understanding of the task. The \textbf{encoding and information representation} specifies how game states, actions, and historical sequences are formatted (\eg, grid layouts for board games, symbolic notations for cards), establishing a consistent interface between the game engine and \acp{llm}. The \textbf{output format requirement} provides strict instructions on how \acp{llm} must structure its response, including required reasoning steps and the final decision, enabling automated parsing and logging of the model's decision process.

The \textbf{user prompt}, delivered at each decision step, supplies \acp{llm} with necessary contextual information to act. \textbf{Player identity information} indicates which role or seat \acp{llm} occupies in the current game. \textbf{State representation} provides a structured textual description of the current game state (\eg, public information, scores). For imperfect-information games, \textbf{game history} offers a sequential record of all publicly observable actions, designed to preserve the Markov property as fully as possible, providing \acp{llm} with sufficient information to reason about the current belief state and infer hidden information. The \textbf{valid action list} enumerates all legal moves available in the current situation, constraining \acp{llm}' choice space and preventing invalid selections.

This two-part prompting framework balances comprehensiveness with flexibility: it provides all essential knowledge and constraints while leaving the reasoning process entirely open to the \acs{llm}'s own strategic analysis. Full prompt templates for each game are provided in \cref{app:prompting}.

\subsection{Selection of Graded Baseline Bots}

The validity of our absolute performance metric hinges on a well-calibrated hierarchy of classic \ac{ai} opponents, sourced from the extensive Botzone library. We construct this hierarchy through a selection process that integrates the Botzone Elo ladder and win-rate analysis against the immediate lower-level baseline. Level 0 (Lv0) is defined as a bot that selects a random legal move, serving as an absolute lower-bound baseline. For each candidate bot at a higher level (Lv1, Lv2, \etc), we require it to demonstrate a clear performance superiority over the preceding tier, with the win rate typically controlled between 70\% and 90\% in head-to-head matches. This range ensures a meaningful and distinguishable skill gap between adjacent levels: neither too narrow to compromise evaluation efficiency, nor too wide to lose discriminative power (\eg, causing all \acp{llm} to fail uniformly against an overwhelmingly strong baseline).

To enhance the robustness and representativeness of each level, we select multiple baseline bots where the Botzone ecosystem allows. Specifically, when the public bot pool is sufficiently large, we include four bots per level; for games with moderate bot availability, two bots are chosen; and for those with limited options, one bot is used. This flexible selection strategy ensures that our benchmark remains widely applicable across games with varying community support.

Furthermore, to ensure our benchmark evaluates general strategic competence rather than specialization against a single algorithmic style, we explicitly prioritize diversity in implementation methods among bots at the same skill tier. Whenever possible, we select multiple bots that employ distinct approaches, such as rule-based, search-based, supervised learning, or reinforcement learning, yet exhibit comparable strength (\eg, similar Elo and win rates against shared baselines). This multi-bot, multi-strategy representation for a given level promotes robustness and reduces evaluation bias, providing a more comprehensive performance anchor against which \acp{llm} can be fairly measured. Detailed descriptions of the selected baseline bots for each game and level are provided in \cref{app:baseline_info}.

\begin{table*}[htbp!]
	\centering
	\setlength{\tabcolsep}{3pt}
	\small
	\resizebox{\linewidth}{!}{%
		\begin{tabular}{@{}llccccccc@{}}
			\toprule
			Model & Tic-Tac-Toe & Gomoku & Ataxx & Reversi & Chess & Texas Hold'em & Landlord & Mahjong \\
			\midrule
			DeepSeek-3.2 & Lv1, 50\% & Lv3, 32.3\% & Lv3, 64.5\% & Lv2, 21.4\% & Lv1, 0.0\% & Lv2, 93.8\% & Lv1, 84.4\% & Lv2, 63.6\% \\
			Claude-Sonnet-4.5 & Lv1, 59.4\% & Lv3, 62.5\% & Lv3, 75.0\% & Lv2, 46.7\% & Lv1, 0.0\% & Lv2, 71.9\% & Lv2, 77.4\% & Lv2, 39.2\% \\
			GPT-5.2 & Lv1, 46.9\% & Lv3, 87.5\% & Lv3, 93.8\% & Lv1, 73.3\% & Lv2, 0.0\% & Lv1, 96.8\% & Lv2, 64.5\% & Lv2, 41.9\% \\
			Gemini3-Pro-Pre.& Lv2, topped & Lv5, 25.0\% & Lv5, 68.8\% & Lv4, 75.0\% & Lv1, 0.0\% & Lv4, topped & Lv4, 50.8\% & Lv3, 89.3\% \\
			Qwen3-235B-Instr. & Lv1, 96.9\% & Lv3, 75.0\% & Lv3, 45.2\% & Lv2, 18.8\% & Lv1, 0.0\% & Lv1, 96.9\% & Lv2, 73.0\% & Lv1, 77.8\% \\
			\midrule
			Qwen3-7B-Instr.& Lv1, 9.4\% & Lv2, 35.7\% & Lv0, 87.5\% & Lv0, 62.5\% & Lv0, 57.1\% & Lv0, 87.5\% & Lv1, 71.9\% & Lv0, 50\% \\
			Qwen3-14B-Instr.& Lv1, 3.1\% & Lv2, 77.4\% & Lv1, 83.9\% & Lv1, 18.8\% & Lv0, 50.0\% & Lv0, 37.5\% & Lv1, 41.3\% & Lv1, 34.8\% \\
			Qwen3-32B-Instr.& Lv1, 21.9\% & Lv3, 45.2\% & Lv3, 58.1\% & Lv1, 32.3\% & Lv1, 0.0\% & Lv1, 63.5\% & Lv1, 88.9\% & Lv1, 42.9\% \\
			\bottomrule
		\end{tabular}%
	}%
	\caption{\textbf{Performance of \acsp{llm} across eight games.} Each cell shows the highest level achieved (Lv0---Lv5) and the win rate against that level. ``Topped'' indicates achieving at least 50\% win rate against the strongest available baseline. Results are based on 100 matches per model-baseline pair (32 per baseline bot when multiple bots exist at the same level). Lv0 represents random play; higher levels indicate progressively stronger traditional \ac{ai} opponents.}
	\label{tab:main_results}
\end{table*}

\subsection{Skill Level Assessment for \acs{llm}-Based Bots}

We assess each \acs{llm}-based bot by assigning it a precise skill rating anchored to our pre-calibrated hierarchy of classic \ac{ai} baselines. This approach provides an absolute, interpretable measure of game-playing ability. The assessment proceeds sequentially from the bottom up. We calculate win rate as $\text{wins}/(\text{wins}+\text{losses}) \times 100\%$, excluding draws. Starting at level 0, a bot advances if it achieves at least 50\% overall win rate against all baselines at its current level. For example, reaching 50\% against Lv0 baselines qualifies it for Lv1; maintaining 50\% against Lv1 baselines then advances it to Lv2, and so forth.

Once we determine the highest level a bot can reach, we express its skill as a (\textit{level}, \textit{progress}) tuple for all games except Tic-Tac-Toe. The \textit{level} indicates the highest baseline tier the bot can match: it has beaten all lower tiers but cannot yet exceed 50\% win rate against this tier's baselines. The \textit{progress} measures how far it has advanced within that tier, calculated as twice its win rate against that tier's baselines. This doubled value represents progress toward the next tier. For instance, a bot with 34\% win rate against Lv1 baselines receives (Lv1, 68\%), meaning it has closed roughly two-thirds of the skill gap between Lv1 and Lv2.

Tic-Tac-Toe presents a special case. Since the game is solved and optimal play always produces draws, win rate is uninformative against perfect opponents. We therefore substitute draw rate for win rate when calculating \textit{progress}.

This (\textit{level}, \textit{progress}) system provides an intuitive, fine-grained, and stable metric for comparing models and tracking performance improvements across training stages.

\section{Experiments and Results}\label{sec:llm_ratings}

Due to time and budget constraints, we evaluated five flagship \acp{llm} representing distinct model families. We tested each with default settings and unlimited output tokens, following the methodology described in \cref{sec:method}. Our selection comprises \textbf{DeepSeek-3.2}, \textbf{Claude-Sonnet-4.5}, \textbf{GPT-5.2}, \textbf{Gemini3-Pro-Preview}, and \textbf{Qwen3-235B-Instruct}. We also evaluated smaller Qwen3 variants (7B, 14B, 32B parameters) to examine how strategic capability scales with model size. Across all models, we completed 6,403 games, collecting 177,047 state-action pairs with accompanying reasoning traces for detailed analysis.

\cref{tab:main_results} summarizes each model's skill ratings across the eight games; complete results appear in \cref{app:performance_results}. An important caveat is that skill levels are comparable only within games, not across them. The number and strength distribution of baseline bots varies substantially between games.

Every model we tested, including the smallest (Qwen3-7B-Instruct), consistently followed game rules and produced properly formatted outputs. This confirms our prompting framework successfully enables \acp{llm} to act as rule-compliant agents, allowing BotzoneBench to focus evaluation on strategic and adaptive capabilities.

All flagship models cleared Lv0 across every game. Among smaller models, performance tracked parameter count: Qwen3-7B-Instruct remained at Lv0 in five games, Qwen3-14B-Instruct in two, while Qwen3-32B-Instruct approached the skill level of the weakest flagship model. This scaling pattern shows that while even small \acp{llm} can follow rules, strategic gameplay depends heavily on model capacity.

Among flagship models, Gemini3-Pro-Preview dominated most games, outperforming competitors by roughly two skill levels on average. Chess was a notable exception: GPT-5.2 led here, consistent with recent specialized studies of \ac{llm} chess understanding~\cite{wen2025chessqaevaluatinglargelanguage}. In Tic-Tac-Toe, every model matched the optimal bot, though with varying consistency. Qwen3-235B-Instruct lost only once across 32 games, while Gemini alone achieved perfect draws in every match.

\section{Case Studies}

This section demonstrates specific capabilities of the BotzoneBench framework through targeted case studies drawn from our evaluation results. \cref{sec:analysis} analyzes \ac{llm} game strategies and behavioral patterns in stochastic, imperfect-information games. In \cref{sec:tournament}, we illustrate how BotzoneBench supports \ac{llm}-vs-\ac{llm} tournaments through a small-scale cross-model competition in Reversi.

\subsection{Performance and Behavior Analysis}\label{sec:analysis}

We analyze behavioral patterns of flagship \acp{llm} in imperfect-information games using both game summaries and interaction logs to understand what drives their performance differences.

\paragraph{Texas Hold'em}
Gemini3-Pro-Preview adopted the most conservative strategy, showing the lowest raise and all-in frequencies and the highest check frequency (\cref{tab:texas_action_rate}). In contrast, GPT-5.2 had the lowest fold rate and highest all-in rate, revealing a more aggressive and ultimately less successful gambling tendency.

\begin{table}[htbp!]
	\centering
	\setlength{\tabcolsep}{3pt}
	\small
	\begin{tabular}{@{}lcccc@{}}
		\toprule
		Model & Fold & Check & All-in & Raise \\
		\midrule
		DeepSeek-3.2 & 27.63 & 55.36 & 1.15 & 31.04 \\
		Claude-Sonnet-4.5 & 34.95 & 63.32 & 0.60 & 19.92 \\
		GPT-5.2 & 23.38 & 66.72 & 2.50 & 18.79 \\
		Gemini3-Pro-Pre. & 30.98 & 74.32 & 0.49 & 11.64 \\
		Qwen3-235B-Instr. & 27.63 & 55.36 & 1.72 & 28.62 \\
		\bottomrule
	\end{tabular}
	\caption{\textbf{Action frequencies (\%) of \acsp{llm} in Texas Hold'em.} Each value represents the percentage of all decisions where the model chose that action.}
	\label{tab:texas_action_rate}
\end{table}

\paragraph{Mahjong}
While other models focused on general Mahjong patterns, Gemini3-Pro-Preview was the only model that explicitly calculated scoring patterns according to MCR rules. This reasoning excerpt from its interaction log demonstrates the capability\footnote{Additional examples from interaction logs are in \cref{app:llm_samples}.}:
\begin{inlinepromptbox}{Fan calculation: Mahjong (Gemini3-Pro-Pre.)}
	\textbf{Case A: Win on B7:} [B4, B4, B4], [B6, B6, B6], [B7, B7, B7], [F2, F2, F2], [W6, W6]. Patterns: All Pungs (6 points), Seat Wind (2 points), and One Voided Suit (1 point) + Pung of Honors (1 point). Total: 10 points - this is a winner...
\end{inlinepromptbox}

Since all evaluated models participated in both Lv0 and Lv1 Mahjong assessments, \cref{tab:fan_stats} aggregates their scoring statistics from these tiers for fair comparison. The table shows Gemini3-Pro-Preview covers the widest variety of patterns, achieving the most distinct patterns valued above 4 points and the highest frequency of games concluding with patterns exceeding 8 points. Gemini's coverage of the widest variety of scoring patterns, combined with interaction logs that reveal its explicit point calculations, suggests a broader and more flexible strategic approach, which helps explain its stronger performance.

\begin{table}[htbp!]
	\centering
	\setlength{\tabcolsep}{3pt}
	\small
	\resizebox{\linewidth}{!}{%
    	\begin{tabular}{@{}lccccc@{}}
    		\toprule
    		& Unique & \multicolumn{2}{c}{Patterns $\geq 8$ points} & \multicolumn{2}{c}{Patterns 4--6 points} \\
    		& patterns & count & freq. & count & freq. \\
    		\midrule
    		DeepSeek-3.2 & 31 & 6 & 9\% & 7 & 22\% \\
    		Claude-Sonnet-4.5 & 30 & 5 & 8\% & 5 & 23\% \\
    		GPT-5.2 & 33 & 5 & 5\% & 7 & 22\% \\
    		Gemini3-Pro-Pre.& 38 & 8 & 13\% & 9 & 19\% \\
    		Qwen3-235B-Instr. & 28 & 4 & 7\% & 5 & 28\% \\
    		\bottomrule
	   \end{tabular}%
    }%
	\caption{\textbf{Scoring pattern diversity of \acp{llm} in Mahjong.} ``Unique patterns'' counts distinct scoring patterns each model achieved across Lv0 and Lv1 games. ``Count'' indicates the number of distinct scoring patterns observed within each category, while ``Freq.'' denotes the frequency of achieved patterns from each category across all winning hands.}
	\label{tab:fan_stats}
\end{table}

\paragraph{Fight the Landlord}
In this game, ``pass'' is always legal, but passing when other options exist is strategic. It can mislead opponents, conserve key cards for later rounds, or set up a stronger play. This transforms the game from simply playing cards to playing minds, turning a straightforward card game into a psychological contest. As \cref{tab:landlord_passing} shows, Qwen3-235B-Instruct almost never passes, while Claude-Sonnet-4.5 frequently passes after its teammate plays:
\begin{inlinepromptbox}{Team coordination: Fight the Landlord (Claude-Sonnet-4.5)}
    I should PASS. The landlord has only 2 cards remaining and already passed on my teammate's pair of Aces. By passing, my teammate maintains control and gets a free play next turn...
\end{inlinepromptbox}

\begin{table}[htbp!]
	\centering
	\setlength{\tabcolsep}{3pt}
	\small
	\resizebox{\linewidth}{!}{%
    	\begin{tabular}{@{}lclc@{}}
    		\toprule
    		Model & Deliberate Pass& Model & Deliberate Pass\\
    		\midrule
    		Qwen3-235B-Instr. & 2.37\% & DeepSeek-3.2 & 10.35\% \\
    		Gemini3-Pro-Pre.& 23.20\% & GPT-5.2 & 22.83\% \\
    		Claude-Sonnet-4.5 & 28.28\% & & \\
    		\bottomrule
    	\end{tabular}%
    }%
	\caption{\textbf{Deliberate passing rates of \acsp{llm} in Fight the Landlord.} Values show the percentage of decisions where the model chose to pass despite having playable cards. Higher rates indicate more strategic passing behavior.}
	\label{tab:landlord_passing}
\end{table}

These distinct passing patterns reflect different behavioral policies in cooperative-competitive play. While such stylistic variation does not predict overall performance, it reveals different approaches to coordinated action and offers valuable insights into strategy emergence in mixed-motive settings.

\subsection{Cross-Model Comparisons}\label{sec:tournament}

Although BotzoneBench primarily uses anchored evaluation against graded \ac{ai} baselines, we retain the capability for direct \ac{llm}-vs-\ac{llm} matches as supplementary validation. Here we present a small-scale cross-model tournament in Reversi using double round-robin format\footnote{The tournament format requires extended playing time, as all moves are predicted by \acp{llm}, and high service stability from both competing models. Due to service instabilities, GPT-5.2 could not complete the full tournament schedule.}. Each model pair played four games, alternating first-mover to control for turn-order bias.

Complete results appear in \cref{tab:reversi_tournament}. While limited in scale, tournament outcomes align with the absolute skill ratings from our anchored evaluation (\cref{tab:main_results}). Notably, Gemini3-Pro-Preview won the most matches and Claude-Sonnet-4.5 the second-most, consistent with their respective skill levels in Reversi. Meanwhile, Qwen3-235B-Instruct and DeepSeek-3.2 ended with even scores against each other, reflecting their comparable ratings in the anchored evaluation.

\begin{table}[htbp!]
	\centering
	\setlength{\tabcolsep}{3pt}
	\small
	\resizebox{\linewidth}{!}{%
    	\begin{tabular}{@{}lcccc@{}}
    		\toprule
    		& DeepSeek & Gemini & Qwen & Claude-Sonnet\\
    		\midrule
    		DeepSeek-3.2 & --- & 0/2 & 0/2 & 1/1 \\
    		Gemini3-Pro-Pre.& 2/0 & --- & 2/0 & 2/0 \\
    		Qwen3-235B-Instr. & 0/2 & 0/2 & --- & 0/2 \\
    		Claude-Sonnet-4.5 & 1/1 & 1/1 & 2/0 & --- \\
    		\bottomrule
    	\end{tabular}%
    }%
	\caption{\textbf{Cross-model tournament results in Reversi.} Each cell shows wins as first mover / wins as second mover. For example, Gemini3-Pro-Preview (row) \vs DeepSeek-3.2 (column) shows 2/0, meaning Gemini won both games where it moved first.}
	\label{tab:reversi_tournament}
\end{table}

\section{Limitations and extensions}

Our framework has two main limitations that suggest directions for future work.

First, we evaluated only five flagship models plus three smaller Qwen3 variants. We excluded Qwen3-235B-Thinking because its extended reasoning chains, which often exceed tens of thousands of tokens in late-game Mahjong, regularly triggered time limits. However, BotzoneBench's modular design readily accommodates additional models and configurations (\eg, temperature, top-k sampling) in future studies.

Second, baseline coverage varies across games. While most games benefit from well-populated bot ecosystems with fine-grained skill ladders, Chess and Texas Hold'em face different challenges. Chess baselines show large performance gaps that reduce discrimination between adjacent skill levels. Texas Hold'em suffers from a shortage of strong public bots, limiting the upper range of our skill hierarchy. Despite these gaps, both games remain valuable for testing deterministic reasoning (Chess) and decision-making under uncertainty (Texas Hold'em). We expect baseline quality to improve as the Botzone community expands.

Beyond addressing these limitations, our framework naturally extends to studying how prompt design affects strategic reasoning. \ac{llm} performance depends heavily on prompt formulation, yet disentangling prompt engineering from inherent reasoning ability remains challenging. Our approach provides game rules and state information without prescriptive strategy hints, but many design choices remain open: level of rule detail, use of heuristics, state encoding format, and access to hidden information. Systematic manipulation of these factors could reveal how much performance stems from prompting versus underlying model capability---a question our rich interaction logs are well-suited to address.

\section{Conclusion}

We introduce BotzoneBench, an evaluation framework that addresses key limitations of existing \ac{llm} game-playing benchmarks. By replacing closed-pool tournaments with skill-calibrated \ac{ai} baselines, we achieve three advantages: absolute performance ratings that remain valid as new models emerge, efficient linear scaling that avoids combinatorial match requirements, and rich interaction data that enables detailed analysis of strategic reasoning patterns.

Our evaluation of eight models across eight games demonstrates the framework's ability to produce fine-grained, interpretable skill assessments while revealing distinct behavioral signatures in how models approach imperfect information and strategic uncertainty. The anchored evaluation methodology generalizes beyond games to any interactive domain with established skill hierarchies, offering a template for systematic, reusable benchmarking.

Future work should expand model coverage, strengthen baseline quality in Chess and Texas Hold'em, and systematically investigate prompt design's role in strategic performance. The interaction logs we collect, 177,047 state-action pairs with reasoning traces, provide a foundation for such studies, enabling researchers to separate prompt engineering effects from core reasoning capabilities.

{
    \bibliographystyle{named}
    \bibliography{reference_header,reference}
}

\clearpage
\crefalias{section}{appendix}
\appendix
\onecolumn
\renewcommand\thefigure{A\arabic{figure}}
\setcounter{figure}{0}
\renewcommand\thetable{A\arabic{table}}
\setcounter{table}{0}
\renewcommand\theequation{A\arabic{equation}}
\setcounter{equation}{0}
\pagenumbering{arabic}
\renewcommand*{\thepage}{A\arabic{page}}
\setcounter{footnote}{0}

\section{Performance Against Baseline Anchors}
\label{app:performance_results}

This appendix provides the complete set of experimental results from which the skill ratings in \cref{sec:llm_ratings} are derived. The performance of each evaluated \ac{llm} is recorded against every level of skill-calibrated baseline bots across all eight games in BotzoneBench.

The results are organized into six tables (\cref{tab:lv0_results,tab:lv1_results,tab:v2_results,tab:v3_results,tab:v4_results,tab:v5_results}), each corresponding to a specific baseline level from Level 0 (random play) to Level 5. Each table cell reports the outcome of an \ac{llm}'s match against a given baseline level for a specific game, using a win-draw-loss notation:
Note that $w-d-l/t$, where $w$ denotes wins, $d$ denotes draws, $l$ denotes losses, and $t$ denotes the total number of completed games. $w + d + l \leq t$, as games that failed to complete after two attempts were discarded and are not counted in the totals.

The tables enable a fine-grained inspection of model performance progression. For instance, they reveal not only the highest level an \ac{llm} can reach (win rate $\ge$50\%), but also its exact win/draw/loss distribution at each intermediary level. This detailed record supports the calculation of the (Level, Progress) ratings presented in \cref{sec:llm_ratings} and allows researchers to trace performance cliffs or consistency patterns (\eg, high draw rates in \textbf{Chess}, or volatile win rates in stochastic games).

Furthermore, these tables serve as a foundational dataset for longitudinal study. As new \acp{llm} are evaluated or as existing models are updated, their results can be added within the same anchored framework, enabling direct and stable comparison over time. The structure ensures that performance is measured against constant, external benchmarks rather than shifting model pools, maintaining the interpretability and scalability central to the BotzoneBench paradigm.

\begin{table*}[htbp]
	\centering
	
	\begin{tabular}{@{}lcccccccc}
		\toprule
		Models/Games    & TicTacToe & Gomoku    & Ataxx     & Reversi   & Chess     & Taxes Hold'em & Landlord   & Mahjong
		\\
		\midrule
		DeepSeek-3.2     & 13-0-3/16 & 16-0-0/16 & 16-0-0/16 & 11-0-5/16 & 16-0-0/16 & 8-0-8/16      & 21-0-11/32 & 11-21-0/32
		\\
		Claude-Sonnet-4.5 & 15-0-1/16 & 16-0-0/16 & 16-0-0/16 & 11-0-5/16 & 16-0-0/16 & 9-0-7/16      & 24-0-8/32  & 15-17-0/32
		\\
		GPT-5.2          & 10-2-2/16 & 16-0-0/16 & 15-0-1/16 & 12-0-4/16 & 15-1-0/16 & 10-0-6/16     & 23-0-9/32  & 14-17-1/32
		\\
		Gemini3-Pro-Pre.     & 15-1-0/16 & 16-0-0/16 & 16-0-0/16 & 16-0-0/16 & 16-0-0/16 & 13-0-3/16     & 32-0-0/32  & 22-10-0/32
		\\
		Qwen3-235B-Instr. & 12-2-2/16 & 16-0-0/16 & 15-0-1/16 & 10-0-6/16 & 16-0-0/16 & 12-0-4/16     & 23-0-9/32  & 12-20-0/32
		\\
		\midrule
		Qwen3-8B-Instr.       & 9-4-3/16  & 14-0-2/16 & 7-0-9/16  & 5-0-11/16 & 2-9-5/16  & 7-0-9/16      & 16-0-16/32 & 1-25-2/32  \\
		Qwen3-14B-Instr.      & 8-2-6/16  & 16-0-0/16 & 13-0-3/16 & 9-0-7/16  & 1-12-2/16 & 3-0-12/16     & 18-0-14/32 & 1-31-0/32  \\
		Qwen3-32B-Instr.      & 16-0-0/16 & 16-0-0/16 & 16-0-0/16 & 10-0-6/16 & 5-11-0/16 & 9-0-7/16      & 17-0-15/32 & 3-27-0/32  \\
		\bottomrule
	\end{tabular}
    \caption{\acp{llm}' Results Against Lv0 Baseline}
	\label{tab:lv0_results}

\end{table*}

\begin{table*}[htbp]
	\centering
	
	\begin{tabular}{@{}lcccccccc}
		\toprule
		Models/Games    & TicTacToe  & Gomoku     & Ataxx      & Reversi    & Chess     & Taxes Hold'em & Landlord   & Mahjong     \\
		\midrule
		DeepSeek-3.2     & 0-16-16/32 & 32-0-0/32  & 24-0-7/32  & 15-1-15/32 & 0-0-32/32 & 34-0-30/64    & 27-0-37/64 & 21-26-15/64 \\
		Claude-Sonnet-4.5 & 0-19-13/32 & 32-0-0/32  & 22-0-10/32 & 20-1-11/32 & 0-0-32/32 & 34-3-27/64    & 43-0-21/64 & 17-34-13/64 \\
		GPT-5.2          & 0-15-17/32 & 31-0-1/32  & 24-0-7/32  & 11-2-19/32 & 3-28-1/32 & 30-2-32/64    & 36-0-27/64 & 24-27-13/64 \\
		Gemini3-Pro-Pre.     & 0-32-0/32  & 32-0-0/32  & 32-0-0/32  & 25-2-4/32  & 0-0-32/32 & 32-4-28/64    & 56-0-8/64  & 30-17-12/64 \\
		Qwen3-235B-Instr. & 0-31-1/32  & 32-0-0/32  & 20-0-12/32 & 18-0-14/32 & 0-0-32/32 & 31-0-33/64    & 43-0-21/64 & 14-28-22/64 \\
		\midrule
		Qwen3-8B-Instr.       & 0-3-29/32  & 21-0-11/32 & N/A        & N/A        & N/A       & N/A           & 23-0-40/64 & N/A         \\
		Qwen3-14B-Instr.      & 0-1-31/32  & 24-0-8/32  & 13-0-19/32 & 3-1-28/32  & N/A       & N/A           & 13-0-50/64 & 4-39-19/64  \\
		Qwen3-32B-Instr.      & 0-7-25/32  & 31-0-1/32  & 17-0-15/32 & 5-0-26/32  & 0-0-32/32 & 20-1-43/64    & 28-0-35/64 & 6-32-22/64  \\
		\bottomrule
	\end{tabular}
    \caption{\acp{llm}' Results Against Lv1 Baseline}
	\label{tab:lv1_results}
\end{table*}

\begin{table*}[htbp]
	\centering
	
	\begin{tabular}{@{}lcccccccc}
		\toprule
		Models/Games    & TicTacToe & Gomoku    & Ataxx      & Reversi   & Chess     & Taxes Hold'em & Landlord   & Mahjong     \\
		\midrule
		DeepSeek-3.2     & N/A       & 25-0-7/32 & 28-0-4/32  & 3-1-28/32 & N/A       & 30-0-34/64    & N/A        & 14-15-30/64 \\
		Claude-Sonnet-4.5 & N/A       & 30-0-2/32 & 20-0-12/32 & 7-2-23/32 & N/A       & 23-0-41/64    & 24-0-38/64 & 10-19-31/64 \\
		GPT-5.2          & N/A       & 24-0-8/32 & 23-0-9/32  & N/A       & 0-0-32/32 & N/A           & 20-0-42/64 & 9-20-34/64  \\
		Gemini3-Pro-Pre.& Topped    & 32-0-0/32 & 32-0-0/32  & 23-0-9/32 & N/A       & 32-0-32/64    & 44-0-20/64 & 38-4-22/64  \\
		Qwen3-235B-Instr. & N/A       & 26-0-6/32 & 24-0-8/32  & 3-0-29/32 & N/A       & N/A           & 23-0-40/64 & N/A         \\
		\midrule
		Qwen3-8B-Instr.       & N/A       & 5-0-23/32 & N/A        & N/A       & N/A       & N/A           & N/A        & N/A
		\\
		Qwen3-14B-Instr.      & N/A       & 12-0-19   & N/A        & N/A       & N/A       & N/A           & N/A        & N/A
		\\
		Qwen3-32B-Instr.      & N/A       & 25-0-6/32 & 17-0-15/32 & N/A       & N/A       & N/A           & N/A        & N/A         \\
		\bottomrule
	\end{tabular}
    \caption{\acp{llm}' Results Against lv2 Baseline}
	\label{tab:v2_results}
\end{table*}

\begin{table*}[htbp!]
	\centering
	
	\begin{tabular}{@{}lcccccccc}
		\toprule
		Models/Games    & TicTacToe & Gomoku     & Gomoku     & Ataxx      & Chess & Taxes Hold'em & Landlord   & Mahjong    \\
		\midrule
		DeepSeek-3.2     & N/A       & 5-0-27/32  & 5-0-27/32  & 10-0-21/32 & N/A   & N/A           & N/A        & N/A        \\
		Claude-Sonnet-4.5 & N/A       & 10-0-22/32 & 10-0-22/32 & 12-0-20/32 & N/A   & N/A           & N/A        & N/A        \\
		GPT-5.2          & N/A       & 14-0-18/32 & 14-0-18/32 & 15-0-17/32 & N/A   & N/A           & N/A        & N/A
		\\
		Gemini3-Pro-Pre.& Topped    & 28-0-3/32  & 28-0-3/32  & 28-0-2/32  & N/A   & 32-2-30/64    & 32-0-32/64 & 25-6-31/64
		\\
		Qwen3-235B-Instr. & N/A       & 12-0-20/32 & 12-0-20/32 & 7-0-24/32  & N/A   & N/A           & N/A        & N/A
		\\
		\midrule
		Qwen3-32B-Instr.      & N/A       & 7-0-25/32  & 7-0-25/32  & 9-0-23/32  & N/A   & N/A           & N/A        & N/A        \\
		\bottomrule
	\end{tabular}
    \caption{\acp{llm}' Results Against Lv3 Baseline}
	\label{tab:v3_results}
\end{table*}

\begin{table*}[htbp!]
	\centering
	
	\begin{tabular}{@{}lcccccccc}
		\toprule
		Models/Games & TicTacToe & Gomoku     & Ataxx      & Reversi    & Chess & Taxes Hold'em & Landlord   & Mahjong \\
		\midrule
		Gemini3-Pro-Pre.& Topped    & 19-0-13/32 & 17-0-14/32 & 12-0-20/32 & N/A   & Topped        & 16-0-47/64 & N/A     \\
		\bottomrule
	\end{tabular}
    \caption{\acp{llm}' Results Against Lv4 Baseline}
	\label{tab:v4_results}
\end{table*}

\begin{table*}[htbp!]
	\centering
	
	\begin{tabular}{@{}lcccccccc}
		\toprule
		Models/Games & TicTacToe & Gomoku    & Ataxx      & Reversi & Chess & Taxes Hold'em & Landlord & Mahjong \\
		\midrule
		Gemini3-Pro-Pre.& Topped    & 4-0-28/32 & 11-0-21/32 & N/A     & N/A   & Topped        & N/A      & N/A     \\
		\bottomrule
	\end{tabular}
    \caption{\acp{llm}' Results Against Lv5 Baseline}
	\label{tab:v5_results}
\end{table*}

\newpage

\section{Baseline Bot Information}
\label{app:baseline_info}
For games with sufficient available bots, we curated baseline bots up to Level 6. In \textbf{Tic-Tac-Toe}, due to the limited strategic depth, Level 1 bots, which consistently outperform random players, are already considered optimal. For \textbf{Texas Hold'em}, we curated only 3 levels due to a lack of significant performance gaps between available candidates. Similarly, for \textbf{Mahjong}, \textbf{Fight the Landlord}, and \textbf{Chess}, we curated 4 levels for the same reasons. \cref{tab:ataxx_levels,tab:fightthelandlord_levels,tab:gomoku_levels,tab:reversi_levels,tab:tictactoe_levels,tab:chess_levels,tab:texasholdem2p_levels,tab:chinesestandardmahjong_levels} provide detailed information on the baseline bots used for each game, including their ranking, score, unique BotID, average rank percentile, and average score.

AVG Rank Percentile(\%) represents the percentage of all ranked bots in the Botzone platform that this baseline bot outperforms. A higher percentage indicates a stronger baseline. For example, An AVG Rank Percentile of 85\% means this baseline is able to win against approximately 85\% of all ranked bots in this game.

\begin{table*}[htbp!]
	\centering
	
	\begin{tabular}{@{}cccccc@{}}
		\toprule
		Level & Rank & Score   & BotID                    & AVG Rank Percentile(\%) & AVG Score \\
		\midrule
		lv1   & 448  & 665.84  & 5af56457433be66904cd5fc9 & 8.71                    & 670.9375  \\
		      & 447  & 668.75  & 609e27f6f827b82eae68c497 &                         &           \\
		      & 446  & 670.99  & 5aefb5aca5858d0880e4b138 &                         &           \\
		      & 445  & 678.17  & 5cef7583947c9c03d0a8856d &                         &           \\
		\midrule
		lv2   & 421  & 770.84  & 5b14092f8472ec612b15720f & 14.96                   & 790.385   \\
		      & 420  & 771.50  & 5b2e9e336a209a1c864fcfc6 &                         &           \\
		      & 418  & 788.88  & 5b13f2aa8472ec612b154262 &                         &           \\
		      & 405  & 830.32  & 5b13842720c3be3c79f942db &                         &           \\
		\midrule
		lv3   & 187  & 1176.64 & 60c33170dce9141a1de3db33 & 62.60245902             & 1184.855  \\
		      & 188  & 1174.37 & 5b0563e1c57fa61d35cf63f5 &                         &           \\
		      & 165  & 1219.50 & 5b12b95c20c3be3c79f8e95c &                         &           \\
		      & 194  & 1168.91 & 60cb300bdce9141a1d06f5a2 &                         &           \\
		\midrule
		lv4   & 3    & 1550.41 & 692679b636184360c6ff49b9 & 98.05327869             & 1522.91   \\
		      & 9    & 1525.23 & 66a1c43238d75d2f025ab988 &                         &           \\
		      & 26   & 1468.63 & 630a0b4266467507cacb6f24 &                         &           \\
		      & 4    & 1547.37 & 61f3c4973e8ab26550d3848b &                         &           \\
		\bottomrule
	\end{tabular}
    \caption{Baseline Information for FightTheLandlord}
	\label{tab:fightthelandlord_levels}
\end{table*}

\begin{table*}[htbp!]
	\centering
	
	\begin{tabular}{@{}cccccc@{}}
		\toprule
		Level & Rank & Score   & BotID                    & AVG Rank Percentile(\%) & AVG Score \\
		\midrule
		lv1   & 4    & 985.80  & 682a9a90d870d24732ac84e4 & 57.14                   & 985.80    \\
		lv2   & 3    & 1105.09 & 675be019f8ef9f3aca87162d & 71.43                   & 1105.09   \\
		lv3   & 2    & 1399.13 & 61d6dcc799f5414277309c51 & 85.71                   & 1399.13   \\
		lv4   & 1    & 1887.08 & 61e9683f3e8ab26550c84f7e & 100.00                  & 1887.08   \\
		\bottomrule
	\end{tabular}
    \caption{Baseline Information for Chess}
	\label{tab:chess_levels}
\end{table*}

\begin{table*}[htbp!]
	\centering
	
	\begin{tabular}{@{}cccccc@{}}
		\toprule
		Level & Rank & Score   & BotID                    & AVG Rank Percentile(\%) & AVG Score \\
		\midrule
		lv1   & 1037 & 376.53  & 586bcf62af944f3f232d615d & 6.80                    & 390.005   \\
		      & 1031 & 406.99  & 5a4841865a2fe258a9abf93e &                         &           \\
		      & 1029 & 390.05  & 5a45d00c5a2fe258a9a5704a &                         &           \\
		      & 1034 & 386.45  & 6808a80ba32ffa0bd4954e73 &                         &           \\
		\midrule
		lv2   & 1021 & 442.45  & 5a6737faf055de4894298bcb & 8.18                    & 461.795   \\
		      & 1020 & 443.54  & 5a51b219fe46681ed44c4006 &                         &           \\
		      & 1016 & 476.85  & 61b5b1278d8bd011d760df70 &                         &           \\
		      & 1013 & 484.34  & 67fcb61543aaa306cc1adb77 &                         &           \\
		\midrule
		lv3   & 999  & 548.54  & 61d013048d8bd011d77ed3b1 & 9.49                    & 523.6975  \\
		      & 1002 & 522.38  & 586529704d074667e2accbfc &                         &           \\
		      & 1003 & 521.65  & 61cfe2348d8bd011d77eaff9 &                         &           \\
		      & 1008 & 502.22  & 61d585198d8bd011d7839426 &                         &           \\
		\midrule
		lv4   & 963  & 672.55  & 5a5dc708fe46681ed45478a5 & 16.24                   & 735.2625  \\
		      & 908  & 764.75  & 680b78ae66c1f905d8bb14dc &                         &           \\
		      & 924  & 748.62  & 5864aaaba932ef1a41c32000 &                         &           \\
		      & 918  & 755.13  & 61d558248d8bd011d7836f15 &                         &           \\
		\midrule
		lv5   & 781  & 895.68  & 680ba82c66c1f905d8bb9ece & 32.75                   & 928.7325  \\
		      & 751  & 921.42  & 5a4092b65a2fe258a9964fad &                         &           \\
		      & 721  & 952.56  & 585f84386d6c3d654a242342 &                         &           \\
		      & 729  & 945.27  & 5a41093d5a2fe258a997caee &                         &           \\
		\midrule
		lv6   & 661  & 1013.18 & 61ab42617b65754c1d4f8477 & 42.07                   & 1029.6675 \\
		      & 640  & 1029.49 & 58627e0d6d6c3d654a24beef &                         &           \\
		      & 635  & 1034.39 & 5a3ca0ea5a2fe258a98cf036 &                         &           \\
		      & 633  & 1041.61 & 5865eb624d074667e2ad0c1b &                         &           \\
		\bottomrule
	\end{tabular}
    \caption{Baseline Information for Ataxx}
	\label{tab:ataxx_levels}
\end{table*}

\begin{table*}[htbp!]
	\centering
	
	\begin{tabular}{@{}cccccc@{}}
		\toprule
		Level & Rank & Score   & BotID                    & AVG Rank Percentile(\%) & AVG Score \\
		\midrule
		lv1   & 42   & 1039.78 & 63cb39e8ee1bce5e6c7f9d12 & 27.27                   & 1040.475  \\
		      & 40   & 1041.17 & 63c684faee1bce5e6c7b68c8 &                         &           \\
		\bottomrule
	\end{tabular}
    \caption{Baseline Information for TicTacToe. Lv1 baseline outperforms the bottom 27.27\% bots and draws with the rest.}
	\label{tab:tictactoe_levels}
\end{table*}

\begin{table*}[htbp!]
	\centering
	
	\begin{tabular}{@{}cccccc@{}}
		\toprule
		Level & Rank & Score   & BotID                    & AVG Rank Percentile(\%) & AVG Score \\
		\midrule
		lv1   & 16   & 958.79  & 63e4f8666ce79f4b2db33ecd & 25.00                   & 958.79    \\
		\midrule
		lv2   & 9    & 1088.64 & 63e384696ce79f4b2db20359 & 62.50                   & 1137.57   \\
		      & 8    & 1186.49 & 63f97abb6ce79f4b2dca6f49 &                         &           \\
		\midrule
		lv3   & 1    & 1501.56 & 63fc5cbe6ce79f4b2dcce80f & 100.00                  & 1501.56   \\
		\bottomrule
	\end{tabular}
    \caption{Baseline Information for TexasHoldem2p}
	\label{tab:texasholdem2p_levels}
\end{table*}

\begin{table*}[htbp!]
	\centering
	
	\begin{tabular}{@{}cccccc@{}}
		\toprule
		Level & Rank & Score   & BotID                    & AVG Rank Percentile(\%) & AVG Score \\
		\midrule
		lv1   & 621  & 876.09  & 5ec4fab8703c1e22d93ec0c0 & 19.76                   & 862.06    \\
		      & 624  & 847.10  & 5eb904a67deb2c02c4a623f2 &                         &           \\
		      & 622  & 863.79  & 5ec0b74c703c1e22d939edcf &                         &           \\
		      & 634  & 861.26  & 5eca3824e9dfff058425851b &                         &           \\
		\midrule
		lv2   & 454  & 962.09  & 5ec3e475703c1e22d93d658a & 40.78                   & 971.65    \\
		      & 474  & 983.97  & 66817f37b54d400b70945271 &                         &           \\
		      & 461  & 984.32  & 5eca8b62e9dfff0584261267 &                         &           \\
		      & 458  & 956.21  & 5eaaddfda96e206b64793d36 &                         &           \\
		\midrule
		lv3   & 242  & 1115.69 & 5fec394cd9383f7579add80f & 70.57                   & 1128.15   \\
		      & 251  & 1136.61 & 694dfd233cb83c184cfd76df &                         &           \\
		      & 218  & 1128.10 & 5fed742dd9383f7579af6982 &                         &           \\
		      & 209  & 1132.21 & 64a11e7ba24a147da2ace510 &                         &           \\
		\midrule
		lv4   & 18   & 1256.04 & 666bc4d8e185683ff83c0cbc & 98.43                   & 1255.96   \\
		      & 8    & 1260.13 & 67aa175c680c9a25f8d53f69 &                         &           \\
		      & 13   & 1254.22 & 684431e6c5819238f6729e70 &                         &           \\
		      & 14   & 1253.44 & 66883144b54d400b70a004fe &                         &           \\
		\bottomrule
	\end{tabular}
    \caption{Baseline Information for Mahjong}
	\label{tab:chinesestandardmahjong_levels}
\end{table*}

\begin{table*}[htbp!]
	\centering
	
	\begin{tabular}{@{}cccccc@{}}
		\toprule
		Level & Rank & Score   & BotID                    & AVG Rank Percentile(\%) & AVG Score \\
		\midrule
		lv1   & 487  & 467.35  & 61cb50238d8bd011d779c378 & 12.47739602             & 490.72    \\
		      & 486  & 475.52  & 61c4d1428d8bd011d7748737 &                         &           \\
		      & 484  & 499.90  & 677b93c2d5b53605c84e2fc7 &                         &           \\
		      & 483  & 520.11  & 6371c8de8d38157ab2ae4d5d &                         &           \\
		\midrule
		lv2   & 462  & 663.46  & 658c4d4926354d2b0ff6b6bd & 16.45569620             & 644.715   \\
		      & 465  & 635.77  & 6591134126354d2b0ffb7286 &                         &           \\
		      & 467  & 609.78  & 5ce4e07ad2337e01c7a6a396 &                         &           \\
		      & 458  & 669.85  & 6598021426354d2b0f01c91e &                         &           \\
		\midrule
		lv3   & 424  & 826.34  & 61c46ae78d8bd011d773ef93 & 22.92043400             & 810.885   \\
		      & 430  & 801.96  & 6578544882ee46246bc5abf5 &                         &           \\
		      & 427  & 814.41  & 6566a5f482ee46246bb5857e &                         &           \\
		      & 428  & 800.83  & 5b1389aa20c3be3c79f94528 &                         &           \\
		\midrule
		lv4   & 325  & 1040.56 & 63bd4cb0ee1bce5e6c7267b5 & 36.75                   & 987.16    \\
		      & 349  & 966.55  & 6593b22b26354d2b0ffdf56d &                         &           \\
		      & 362  & 966.30  & 5b0a17975de6fd5a623b683f &                         &           \\
		      & 367  & 975.23  & 5b113d80b10ac8057afddc1b &                         &           \\
		\midrule
		lv5   & 154  & 1330.64 & 5b51210cb69487015e06f615 & 71.88                   & 1329.02   \\
		      & 156  & 1329.79 & 658d130726354d2b0ff78fff &                         &           \\
		      & 157  & 1329.26 & 6185f8c5dc5aa23c13afa1f1 &                         &           \\
		      & 159  & 1326.39 & 61cc812c8d8bd011d77ac110 &                         &           \\
		\midrule
		lv6   & 31   & 1615.62 & 599b9ce3579ce9447f9f5843 & 95.61                   & 1643.945  \\
		      & 26   & 1638.42 & 594c60df9f223932cc47841a &                         &           \\
		      & 23   & 1657.76 & 5b51c949b69487015e074936 &                         &           \\
		      & 21   & 1663.98 & 5ae5f3ec4623c5389913a399 &                         &           \\
		\bottomrule
	\end{tabular}
    \caption{Baseline Information for Gomoku}
	\label{tab:gomoku_levels}
\end{table*}

\begin{table*}[htbp!]
	\centering
	
	\begin{tabular}{@{}cccccc@{}}
		\toprule
		Level & Rank & Score   & BotID                    & AVG Rank Percentile(\%) & AVG Score \\
		\midrule
		lv1   & 601  & 646.58  & 5677b2bd41c47e1a0f3622ad & 17.78                   & 655.1025  \\
		      & 599  & 657.99  & 5c8906d683652225021eca3e &                         &           \\
		      & 598  & 660.78  & 567ad752e98e37a623ef1d9d &                         &           \\
		      & 597  & 655.06  & 5fe35b71d9383f7579a4c2d2 &                         &           \\
		\midrule
		lv2   & 548  & 781.10  & 568e729f0e303a823144533b & 25.10                   & 784.9575  \\
		      & 541  & 788.77  & 5685fc629aab81be75ae57d4 &                         &           \\
		      & 546  & 781.46  & 5b23253f0edf94798cdc3d3e &                         &           \\
		      & 547  & 788.50  & 65a2ac235652b005b946e542 &                         &           \\
		\midrule
		lv3   & 472  & 885.11  & 568d41940e303a82314441ab & 34.00962861             & 872.855   \\
		      & 481  & 871.58  & 6583232282ee46246bce44e7 &                         &           \\
		      & 492  & 857.83  & 65a4194f5652b005b94906b5 &                         &           \\
		      & 478  & 876.90  & 65a15bff26354d2b0f0a883e &                         &           \\
		\midrule
		lv4   & 361  & 1061.55 & 5a5a1077fe46681ed451ecf1 & 49.95                   & 1055.935  \\
		      & 363  & 1057.95 & 6582f63982ee46246bce2dfa &                         &           \\
		      & 366  & 1054.61 & 658d836726354d2b0ff7deda &                         &           \\
		      & 369  & 1049.63 & 6590341526354d2b0ffa7d4b &                         &           \\
		\midrule
		lv5   & 271  & 1176.90 & 5e740a94e952081b8838758a & 62.83                   & 1178.7075 \\
		      & 275  & 1168.57 & 5a5a4753fe46681ed4521299 &                         &           \\
		      & 270  & 1184.42 & 5a5b50d3fe46681ed452cb00 &                         &           \\
		      & 269  & 1184.94 & 659ff01526354d2b0f093aea &                         &           \\
		\midrule
		lv6   & 151  & 1367.78 & 65a1465526354d2b0f0a7424 & 82.08                   & 1432.12   \\
		      & 121  & 1467.10 & 5c91cbf19f425613e1cf1b26 &                         &           \\
		      & 126  & 1447.20 & 68eb5c80dcee294b521c5162 &                         &           \\
		      & 127  & 1446.40 & 5ac1075cc5d5e3196fc6ea48 &                         &           \\
		\bottomrule
	\end{tabular}
    \caption{Baseline Information for Reversi}
	\label{tab:reversi_levels}
\end{table*}

\newpage
\section{Prompting for \acsp{llm}}
\label{app:prompting}
This section details the prompts used to evaluate \acp{llm} across various games. Each prompt is divided into two parts. \textbf{System Prompt} contains static instructions, including game rules, strategic guidelines, and the required response format. \textbf{User Prompt} contains variable game state information (\eg, board positions, player hands, valid moves), which is dynamically converted to text. Variable elements are indicated by \underline{underlines}, and repetitive sequences are abbreviated with "..." to save space. The prompts were designed to instruct the \ac{llm} to act as a player, analyze the current state, and output a valid move in a specified format, preceded by a brief strategic evaluation.

\subsection{Ataxx}
\begin{promptbox}{Prompt for Ataxx}
	\subsubsection*{System Prompt}
	You are an Ataxx game player. \\
	Ataxx is a two-player board game played on a 7x7 grid, where players take turns moving pieces. \\
	Each turn, a piece can either clone to an adjacent square or jump to a square two spaces away. \\
	After placing a piece, all adjacent opponent pieces are flipped. \\
	If there are no legal moves, the turn is skipped. \\
	Please check the current board state and provide your move legally. Unless there is truly no legal move, always choose a move. \\
	The board uses A-G / 1-7 coordinates (Pay attention to the Range!).

	\medskip
	Important instructions: Please output your answer directly in your reply without calling any tool functions.\\
	Provide your analysis and suggested move in plain text.\\
	Now is your turn, to your best ability, please briefly evaluate a few promising moves, provide reason for your final choice, and output the action with specified format.

	\medskip
	Your response format should be: \\
	Evaluation: [Your Evaluation] \\
	Reason: [Your Reason] \\
	Answer: [Your Answer] \\
	Output valid action if possible, the coordinate should be in the form of (A,1) -> (B,2).\\
	If there is no valid action, output 'None'.

	\medskip
	\subsubsection*{User Prompt}
	Your pieces are marked with \underline{'@'}, opponents' are marked with \underline{'*'}, and empty spaces are '.' \\
	This is your board state representation.

	\smallskip
	\begin{tabular}{|c|c|c|c|c|c|c|c|}
		\hline
		  & A & B & C & D & E & F & G \\ \hline
		1 & @ & . & . & . & . & . & * \\ \hline
	\end{tabular}
	\smallskip \\
	...\\
	\begin{tabular}{|c|c|c|c|c|c|c|c|}
		\hline
		7 & * & . & . & . & . & . & @ \\ \hline
	\end{tabular}

	\medskip
	Your valid actions are: \uline{(A, 1) -> (A, 2), (B, 1) -> (C, 3),...}	\\
	"Your previous invalid outputs are:\underline{”…”}
\end{promptbox}

\subsection{Chess}
\begin{promptbox}{Prompt for Chess}
	\subsubsection*{System Prompt}
	You are a Chess player. Chess is a 2-player game playing on an 8x8 board, and the goal is to checkmate your opponent's king.

	\medskip
	\textbf{Pieces \& Movement:} \\
	* King (K): Moves one square in any direction. \\
	* Queen (Q): Moves any distance in any straight or diagonal line. \\
	* Rook (R): Moves any distance horizontally or vertically. \\
	* Bishop (B): Moves any distance diagonally. \\
	* Knight (N): Moves in "L" shape: 2 squares one way + 1 square perpendicular. \\
	* Pawn (P): \\
	\indent - Forward 1 square (never backward) \\
	\indent - First move: optionally 2 squares forward \\
	\indent - Captures diagonally forward \\
	\indent - Promotion: Reaches last rank → any piece (except king)

	\medskip
	\textbf{Special rules:} \\
	* Castling: King moves 2 squares toward rook, rook jumps over. Conditions: \\
	\indent - Neither piece has moved \\
	\indent - No pieces between \\
	\indent - King not in check \\
	\indent - King doesn't pass through attacked squares \\
	* En Passant: Pawn that moved 2 squares can be captured by adjacent enemy pawn as if it moved 1 square. \\
	* Check: King under attack must escape immediately. \\
	* Checkmate: King in check with no legal moves → game ends. \\
	* Stalemate: King not in check but has no legal moves → draw.

	\medskip
	\textbf{Board and Chess piece Representation:} \\
	* Board is represented as [a-h][1-8] \\
	* Origin: bottom-left corner (a1) \\
	* White side: Bottom (ranks 1-2) \\
	* Black side: Top (ranks 7-8) \\
	* X-axis: Files \texttt{a} to \texttt{h} (left to right) \\
	* Y-axis: Ranks \texttt{1} to \texttt{8} (bottom to top)

	\medskip
	\textbf{Piece encoding format:} [Type][Color][2Step][Moved] \\
	* Type (Piece Type): \\
	\indent - \texttt{K} - King \\
	\indent - \texttt{Q} - Queen \\
	\indent - \texttt{R} - Rook \\
	\indent - \texttt{B} - Bishop \\
	\indent - \texttt{N} - Knight \\
	\indent - \texttt{P} - Pawn \\
	* Color (Player): \\
	\indent - \texttt{W} - White (moves first) \\
	\indent - \texttt{B} - Black \\
	* 2Step (En Passant flag): \\
	\indent - \texttt{F} - Did NOT move 2 squares last turn \\
	\indent - \texttt{T} - Moved 2 squares last turn (en passant eligible) \\
	* Moved (Movement flag): \\
	\indent - \texttt{1} - Has moved before \\
	\indent - \texttt{0} - Has NOT moved yet

	\medskip
	\textbf{Examples:} \\
	* \texttt{PWF1}: White Pawn, not available for en passant, already moved \\
	* \texttt{RBF0}: Black Rook, never moved (castling eligible) \\
	* \texttt{KWT1}: White King, irrelevant 2-step flag, has moved \\
	* \texttt{QWF0}: White Queen, irrelevant 2-step flag, never moved

	\medskip
	Now it is your turn. Choose your move to maximize your chances of winning. \\
	Please briefly evaluate a few promising moves, provide the reason for your final choice, and output one of the valid actions with specified format.\\
	Your response format should be: \\
	Evaluation: [Your Evaluation] \\
	Reason: [Your Reason] \\
	Answer: [Your Answer]

	\medskip
	\subsubsection*{User Prompt}
	You are playing as White. \\
	Current state representation is as follows:

	\smallskip
	\begin{tabular}{|c|c|c|c|c|c|c|c|c|}
		\hline
		  & a    & b    & c    & d    & e    & f    & g    & h    \\ \hline
		8 & RBF1 & NBF1 & BBF1 & QBF1 & KBF1 & BBF1 & NBF1 & RBF1 \\ \hline
	\end{tabular}
	\smallskip \\
	...\\
	\begin{tabular}{|c|c|c|c|c|c|c|c|c|}
		\hline
		1 & RWF1 & NWF1 & BWF1 & QWF1 & KWF1 & BWF1 & NWF1 & RWF1 \\ \hline
	\end{tabular}

	Your valid moves are (src-dst-promotion, or src-dst): \\
	\uline{[P: a2-a3, P: a2-a4, P: b2-b3, ..., N: g1-h3]}

	Your answer format should be: Answer: piece-name: move.\\
	Piece-name and move (src-dst-promotion or src-dst) should be directly copied EXACTLY from valid move list.\\
	Example of valid output format: N: a1-b3; P: e7-e8-q; Piece B: g5-f4.\\
	Please select a move from your valid moves that maximize your probability of winning.

\end{promptbox}

\subsection{Gomoku}
\begin{promptbox}{Prompt for Gomoku}
	\subsubsection*{System Prompt}
	You are a Gomoku (Five-in-a-Row) game player. \\
	Gomoku is a 15x15 board game. Columns are labeled A-O. Rows are labeled 1-15. \\
	The goal of Gomoku is to connect 5 stones in any row, column, or diagonal lines to win. \\
	Opponents may block your attempts by intercepting your lines.

	\medskip
	\textbf{General Strategies:} \\
	1. \textbf{Board Survey:} Identify 'Hot Zones' where stones are clustered. Where is the momentum? \\
	2. \textbf{Threat Assessment:} Does the opponent have any immediate lethal threats (4-in-a-row or open-3) in any rows, columns, or diagonal lines? \\
	3. \textbf{Winning Moves:} Can you create a 4-in-a-row or open-3 in any row/col/diag? \\
	4. \textbf{Winning:} If you have any locations to immediately form a 5-in-a-row, you should play that position to win.

	\medskip
	Prepare a valid JSON object like \texttt{\{"col": "H", "row": 8\}} for your answer, along with your explanation and reason. \\
	Now it is your turn. Choose your move to maximize your chances of winning. Briefly evaluate a few promising moves, provide the reason for your final choice, and output the action with specified format.

	\medskip
	Your response format should be: \\
	Evaluation: [Your Evaluation] \\
	Reason: [Your Reason] \\
	Answer: [Your Answer in JSON]

	\medskip
	\subsubsection*{User Prompt}
	You are playing as \uline{White (O)}. It is your turn to move. \\
	Board 15x15:

	\smallskip
	\begin{tabular}{|c|*{15}{c|}}
		\hline
		  & A & B & C & D & E & F & G & H & I & J & K & L & M & N & O \\ \hline
		1 & . & X & . & O & . & . & . & . & . & . & . & . & . & . & . \\ \hline
	\end{tabular}
	\smallskip \\
	... \\
	\begin{tabular}{|c|*{15}{c|}}
		\hline
		15 & . & . & . & . & X & . & . & . & . & . & . & . & . & . & . \\ \hline
	\end{tabular}

	Occupied Coordinates: \uline{B2, D4, H8}

	Analyze the board above. \\
	Find a winning move that is legally valid (Is it in the 'Occupied' list?).\\
	Remember to provide Explanation, Reason, and Answer.\\
	\uline{Invalid format. Please output strictly JSON: \{"col": "Letter", "row": Number\}}
\end{promptbox}

\subsection{Landlord}
\begin{promptbox}{Prompt for Fight The Landlord}
	\subsubsection*{System Prompt}

	You are a Fight The Landlord (Dou Dizhu) player.

	\medskip
	\textbf{Introduction:} \\
	Fight the Landlord is a poker climbing game primarily for three players.

	\medskip
	\textbf{Players, Cards and Deal:} \\
	This game uses a 54-card pack including two jokers, red and black. The cards rank from high to low: red joker, black joker, 2, A, K, Q, J, 10, 9, 8, 7, 6, 5, 4, 3. Suits are irrelevant.

	\medskip
	\textbf{Play:} \\
	The landlord plays first, and may play a single card or any legal combination. Each subsequent player in anticlockwise order must either pass (play no card) or beat the previous play by playing a higher combination of the same number of cards and same type. Exceptions: a rocket beats any combination; a bomb beats any combination except a higher bomb or rocket.

	\medskip
	\textbf{Combinations:} \\
	1. Single card: ranking from three (low) up to red joker (high). \\
	2. Pair: two cards of the same rank. \\
	3. Triplet: three cards of the same rank. \\
	4. Triplet with an attached card: a triplet + any single card. \\
	5. Triplet with an attached pair: a triplet + a pair. \\
	6. Sequence: at least five consecutive cards, 3-A (2/jokers excluded). \\
	7. Sequence of pairs: at least three consecutive pairs, 3-A. \\
	8. Sequence of triplets: at least two consecutive triplets, 3-A. \\
	9. Sequence of triplets with attached cards: each triplet + one extra card. \\
	10. Sequence of triplets with attached pairs: each triplet + one extra pair. \\
	11. Bomb: four cards of the same rank. Beats everything except a rocket or higher bomb. \\
	12. Rocket: a pair of jokers. The highest combination, beats everything. \\
	13. Quadplex set: a bomb + two single cards OR two pairs.

	Use the EXACT card symbols provided in the Valid Moves list.

	Prepare a valid JSON object like
	\texttt{"cards": ["\ensuremath{\spadesuit}3", "\ensuremath{\heartsuit}3"]}
	for your answer, along with your explanation and reason. If you pass, output: {"cards": []}\\

	Now is your turn, to your best ability, choose your move to maximize your chances of winning.\\
	Please briefly evaluate a few proimsing moves, provide reason for your final choice, and output the action with specified format.\\

	\medskip
	Your response format should be: \\
	Evaluation: [Your Evaluation] \\
	Reason: [Your Reason] \\
	Answer: [Your Answer in JSON]

	\medskip
	\subsubsection*{User Prompt}
	You are playing as \uline{a peasant. Your aim is to play out all your cards before the landlord does, OR help your teammate (the other peasant) play out their cards before the landlord does.}

	\uline{You are a Peasant. The Previous Player (Left) is the Landlord (ENEMY). The Next Player (Right) is your TEAMMATE (Peasant). You must cooperate with your teammate to beat the Landlord while trying to empty your own hand.}

	\medskip
	Your Hand:\uline{ \texttt{["\ensuremath{\spadesuit}3"}, ..., "\ensuremath{\diamondsuit}2", "BlackJoker"]} \\
	Last Play to Beat: \uline{\texttt{["\ensuremath{\clubsuit}J", "\ensuremath{\diamondsuit}J"]}} \\

	\medskip
	--- Game History (Recent) ---: \\
	\uline{
		Turn 1 (Next Player): \ensuremath{\spadesuit}3 \\
		Turn 1 (Prev Player): Pass \\
		...\\
		Turn 3 (Prev Player): \ensuremath{\clubsuit}{J} \ensuremath{\diamondsuit}{J} }\\

	\medskip
	--- Player Info ---: \\
	Cards Held by Next Player (Right) [\uline{Peasant (Teammate)}]: \uline{12}\\
	Cards Held by Prev Player (Left) [\uline{Landlord (Opponent)}]: \uline{15}

	\medskip
	--- Cards Remaining Unknown (Card Counting) ---:\\
	(Format explanation: 'Rank: Count', \eg, '2: 4' means there are four 2s left unknown)\\
	\uline{RedJoker: 1, BlackJoker: 0, ..., 3: 4}

	\medskip
	Valid Moves: \\
	\uline{
	\texttt{"cards": ["\ensuremath{\spadesuit}3"]} \\
	...\\
	\texttt{"cards": ["\ensuremath{\clubsuit}9", "\ensuremath{\diamondsuit}9", "\ensuremath{\spadesuit}9", "\ensuremath{\spadesuit}10", "\ensuremath{\spadesuit}10"]} \\
	rocket: [BlackJoker, RedJoker] \\
	pass: []}

	\medskip
	Select the best move from the Valid Moves list above. You can choose to pass even if you have valid moves that can beat the previous play, but you cannot pass if you are in a Free Play (there is no last play to beat).
	Ensure your response matches one of the valid moves exactly.
	Remember to provide Explanation, Reason, and Answer.

	\medskip
	\uline{Invalid format. Please output strictly JSON: {"cards": [...]}}

\end{promptbox}


\subsection{Mahjong}
\begin{promptbox}{Prompt for Mahjong}
	\subsubsection*{System Prompt}
	You are a Chinese Standard Mahjong (MCR Mahjong) player. It is a four-player game, and it is played with a basic set of 136 tiles with Chinese characters and symbols. \\

	Players draw and discard tiles in turn until they complete a winning hand with a 14th tile. The basic type of winning hand consists of four melds and a pair, while there exist winning hands with s[...] \\

	\textbf{Tiles and Representation:}  \\
	- There are 4 identical copies of each suited tile and honored tiles.  \\
	- Tiles are represented using Letter+Number, i.e. W7, T2, B4, F1, J3, etc.  \\
	- Suited tiles are divided into 3 suits: Characters, Bamboos, and Dots, each numbered from 1 to 9.  \\
	- Letter 'W' represents Character suit,  'T' for Bamboo suit, 'B' for Dots, \eg, W3 is Character-3, T8 is Bamboo-8, B5 is Dot-5.  \\
	- Honor tiles are divided into two sets: Wind tiles of four directions and Dragon tiles of three colors.  \\
	- Letter 'F' represents Wind tiles, specifically, 'F1' is East Wind, 'F2' is South Wind, 'F3' is West Wind, 'F4' is North Wind.  \\
	- Letter 'J' represents Dragon tiles, specifically, 'J1' is Red Dragon, 'J2' is Green Dragon, 'J3' is White Dragon.  \\

	\textbf{Melds:}  \\
	Melds are groups of tiles within players' hands, which are essential components to form a winning hand. There are three kinds of melds: Pung, Kong, and Chow.\\
	- A Pung is three identical tiles (suited or honor). Action 'Peng' forms a Pung from a just-discarded tile.  \\
	- A Kong is a set of four identical tiles. While it occupies four physical tiles, it counts as a completed meld (like a Pung) for hand formation and scores additional points. Action 'Gang' means f[...]. \\
	- A Chow is three suited tiles of the same suit in a consecutive numerical sequence. Action 'Chi' means forming a Chow using a tile just discarded by the player immediately before you (your left-h[...]\\
	'Chi Tile' maybe contained in valid actions or histories. It means that a player uses two tiles from his hand tiles, and a tile just discards to form a Chow [Tile-1, Tile, Tile+1].\\

	\textbf{Important Note on 'Chi' Action Notation:}  \\
	In the context of valid actions or game history, you may encounter a notation like 'Chi W2'. \\
	This is a **compact representation** meaning that the player can perform or have performed a 'Chi' action to complete the specific Chow sequence **[W1, W2, W3]**. \\
	Crucially, the tile they actually claimed from the previous player's discard could have been **any one of the three tiles** in that sequence (W1, W2, or W3), depending on which tile completed the [...]\\
	The notation 'Chi X' (\eg, 'Chi W2') always identifies the **completed Chow sequence** (\eg, [W1, W2, W3]), not the discarded tile taken. \\

	Chinese Standard Mahjong specifies 80 different scoring patterns, each worth some points, and a player can only win when their hand is worth no less than 8 points.\\
	Below are the explanation and examples for scoring patterns.\\

	\textbf{Explanation and Examples for Scoring Patterns:}  \\
	\textbf{Patterns worth 88 points:}  \\
	- Big Four Winds: pungs/kongs of all four winds. Example: [F1, F1, F1, F2, F2, F2, F3, F3, F3, F4, F4, F4, B3, B3]  \\
	... \\

	\textbf{Worth 64 points:}  \\
	- All Terminals:A hand consisting of only terminal tiles, i.e., number 1 or 9 of each suit. Example: [T1, T1, T1, B1, B1, B1, W9, W9, W9, B9, B9, B9, T9, T9]  \\
	...\\

	...\\

	\textbf{Worth 1 point:}\\
	- Pure Double Chow: A hand with two identical chows. This pattern can be counted more than once if there are multiple melds meeting this requirement.\\
	...\\

	\medskip
	Now is your turn, to your best ability, please briefly evaluate a few proimsing moves, provide reason for your final choice, and output one of the valid actions with specified format, such as Play[...]\\
	Your response format should be exactly:\\
	Evaluation: [Your Evaluation]\\
	Reason: [Your Reason]\\
	Answer: [Your Answer]\\

	\subsubsection*{User Prompt}
	The following is the state information: \uline{The Prevailing Wind is: East Wind. Your Seat Wind is: South Wind.}  \\
	\uline{Unshown tiles are: [W1: 3; W2: 4; ...; J3: 4].\\
	Your hand tiles are: [W1: 2; W3: 1; ...; J1: 2].\\
	Your melds are: [[W3, W3, W3]].\\
	The melds of the next player: [[T5, T5, T5]].\\
	The melds of the player sitting across: [[]].\\
	The melds of the previous player: [[B4, B5, B6]].\\
	Your game history is: Play W5, Play T3.\\
	The game history of the next player is: Play B7, Peng T5.\\
	The game history of the player sitting across is: Play W2.\\
	The game history of the previous player is: Play T9, Gang B3.}\\
	The followings are your valid actions:\uline{[Play W3; Play T9; BuGang B7; Pass]}\\
	\uline{Your previous invalid answers are: I think you should Play W5 because it is safe, Hu looks good too.}

\end{promptbox}

\subsection{Reversi}
\begin{promptbox}{Prompt for Reversi}
	\subsubsection*{System Prompt}
	Reversi (also known as Othello) is a two-player board game where players take turns placing pieces on an 8x8 board.

	On each turn, a player may place a piece only in a position that meets both of the following conditions:\\

	1. The position is empty. \\
	2. From that position, in at least one of the eight directions (horizontal, vertical, or diagonal), there is a piece of the player's own color such that all positions between it and the new piece ar[...]\\

	After placing the piece, the player must flip all of the opponent's pieces that are sandwiched between their new piece and any of their existing pieces in any direction. \\
	If a player has no valid move available, their turn is skipped. \\
	The game ends when both players have no valid moves. The player with the greater number of pieces on the board wins.\\

	The board is represented using coordinates A-H for columns and 1-8 for rows. \\
	You are a Reversi player, and it is currently your turn. \\
	Choose your move to maximize your chances of winning.\\

	Now is your turn, to your best ability, please briefly evaluate a few proimsing moves, provide reason for your final choice, and output the action with specified format.\\
	Your response format should be:\\
	Evaluation: [Your Evaluation]\\
	Reason: [Your Reason]\\
	Answer: [Your Answer]\\
	Output valid action if possible, the coordinate should be in the form of (A-H,1-8).\\
	If there is no valid action, output 'None'

	\medskip
	\subsubsection*{User Prompt}
	Your piece will be marked with \uline{'*'}, opponent's with \uline{'@'}, and empty space with '.'.

	This is your game state:

	\smallskip
	\begin{tabular}{|c|c|c|c|c|c|c|c|c|}
		\hline
		  & 1 & 2 & 3 & 4 & 5 & 6 & 7 & 8 \\ \hline
		A & . & . & . & * & @ & . & . & . \\ \hline
	\end{tabular}
	\smallskip \\
	...\\
	\begin{tabular}{|c|c|c|c|c|c|c|c|c|}
		\hline
		H & . & . & . & @ & * & . & . & . \\ \hline
	\end{tabular}

	Your valid actions are: \uline{(C, 4), (D, 3), (E, 6), (F, 5)}

	\uline{Your previous invalid outputs: (D,3), (No Valid Move)}
\end{promptbox}

\subsection{Texas Hold'em}
\begin{promptbox}{Prompt for Texas Hold'em}
	\subsubsection*{System Prompt}
	You are a 2-player Texas Hold'em Player. \\
	The game's objective is to make the best 5-card poker hand using any combination of your 2 private cards and the 5 community cards.\\

	Rounds in Texas Hold'em:
	\begin{enumerate}
		\item Pre-flop: Each player gets 2 face-down cards. Betting begins.
		\item The Flop: 3 community cards dealt face-up. Another betting round.
		\item The Turn: A 4th community card dealt face-up. Betting round.
		\item The River: The 5th and final community card dealt face-up. Final betting round.
		\item Showdown: Remaining players reveal their hands. The best hand wins the pot.\\
	\end{enumerate}

	Hand Rankings (High to Low): Royal Flush, Straight Flush, Four of a Kind, Full House, Flush, Straight, Three of a Kind, Two Pair, One Pair, High Card.\\
	The hand will be explained in detail below:\\

	\textbf{Note for Action representation:}
	\begin{enumerate}
		\item Royal Flush: The highest possible hand: A, K, Q, J, 10, all the same suit. Example: Your hand: A of Spades, K of Spades | Board: Q of Spades, J of Spades, 10 of Spades, 5 of Diamonds, 2 of Hearts, [...]
		\item ... \\
		\item One Pair: Two cards of the same rank. Example: K of Hearts, K of Spades, Q of Diamonds, 10 of Clubs, 8 of Hearts.
		\item High Card: No made hand; highest single card wins. \\
	\end{enumerate}
	Note for Action representation: -1 for Fold, -2 for All-in. 0 for Call/Check. >0 for the amount you raise the bet. \\

	When placing a raise, the chips bet must be at least twice the current maximum bet of this round. If no player has bet before in this round, the raise must be at least equal to the big blind.\\

	To call or raise, a player must have enough chips to cover the required bet. Otherwise, they can only fold or go all-in. An improper bet is considered an illegal action and will be treated as a fold.\\

	Now is your turn, to your best ability, please select an action from your valid actions that maximize your probability of winning. If you have more chips than your opponent by the last hand of the game, play accordingly to maximize final victory (\eg, more risk-averse or risk-seeking as appropriate).\\
	Please briefly evaluate a few promising moves, provide reason for your final choice, and output the action with specified format.\\
	Your response format should be:\\
	Evaluation: [Your Evaluation]\\
	Reason: [Your Reason]\\
	Answer: [Your Answer] (should be a number)\\

	\medskip
	\subsubsection*{User Prompt}
	This is the \uline{1} hand out of \uline{20} hands. Everyone starts with 20,000 chips each hand, and you have \uline{19950}. Your cumulative win/loss on chips until last hand are \uline{0}, but the cumulative win/loss chips are inaccessible and independent from your current chips (consider them as deposits/debts to a bank). \uline{You are currently losing to your opponent}.\\
	Your cards are: \uline{Ace of Hearts, 8 of Spades}.\\
	\uline{Currently there is no public card shown}.\\
	The Game history is as follows: \\
	\uline{Your valid actions are: [-2 (All-in), -1 (Fold), 0 (Check), 200 (raise the bet), or any value above 200 but lower than your current chips.].}

\end{promptbox}

\subsection{TicTacToe}
\begin{promptbox}{Prompt for TicTacToe}
	\subsubsection*{System Prompt}
	You are a TicTacToe expert. You and your opponent are playing as 'X' or 'O', on a 3x3 board.

	Columns are labeled from A-C. Rows are labeled from 1-3.

	You want to win or draw by forming a 3-in-a-row in any rows, columns, or diagonal lines.

	Please output a valid JSON object like \{"col": 'Letter', "row": Number\} for your answer.

	Please briefly evaluate a few promising moves, provide reason for your final choice, and output the action with specified format.

	Your response format should be: \\
	Evaluation: [Your Evaluation] \\
	Reason: [Your Reason] \\
	Answer: [Your Answer in JSON]

	\medskip
	\subsubsection*{User Prompt}
	It's your turn to move and you are playing as \uline{X}.

	Your board representation is: \\
	\smallskip
	\begin{tabular}{|c|c|c|c|}
		\hline
		  & A & B & C \\ \hline
		1 & X & O & . \\ \hline
		2 & X & . & . \\ \hline
		3 & . & . & O \\ \hline
	\end{tabular}

	The occupied coordinates are: \uline{A1, A2, B1, C3}. \\
	The valid moves are: \uline{A3, B2, B3, C1, C2}.
\end{promptbox}

\section{Samples of \texorpdfstring{\acsp{llm}}' Reasoning Traces}\label{app:llm_samples}

In this appendix, we provide representative reasoning traces from \acp{llm} to substantiate the arguments presented in \cref{sec:llm_ratings}.

\begin{promptbox}{Team Coordination 1: Fight The Landlord (Gemini3-Pro-Preview)}

	Reason: Teammate is close to winning (3 cards left) and played a low single (9). Beating a teammate's low single is generally a bad move as it obstructs their potential plan to regain initiative or discard small cards. I should conserve my strength (the single 2 and the long straight) for when the Landlord plays or to take control later if necessary. Passing allows the pressure to remain on the Landlord.

	\medskip
	Answer: Pass
\end{promptbox}

\begin{promptbox}{Team Coordination 2: Fight The Landlord (Gemini3-Pro-Preview)}
	You should not beat your teammate\'s high card (King) unless you can surely win or force a specific necessary outcome. By passing, you force the Landlord (who has only 7 cards left) to spend a high card (Ace, 2, or Joker) to beat the King. If the Landlord passes, your teammate maintains control. Saving your Ace and 2s to catch the Landlord later is the optimal strategy

	\medskip
	Answer: Pass
\end{promptbox}

\begin{promptbox}{Team Coordination: Fight The Landlord (Claude-Sonnet-4.5)}
	Playing \ensuremath{\diamondsuit}J would be the cheapest option if I wanted to play, but there's no strategic benefit to overriding my teammate's control. My teammate just successfully beat the Landlord's \ensuremath{\diamondsuit}9 with \ensuremath{\spadesuit}10.

	\medskip
	Answer: Pass
\end{promptbox}

\begin{promptbox}{Secure Advantage: Fight The Landlord (Gemini3-Pro-Preview)}
	Should I use the \ensuremath{\clubsuit}2 or the Black Joker? The Joker beats everything except the Red Joker, but if the Landlord has the 2, I'm in trouble. ... If I play the Joker, the Landlord can beat it with the Red Joker, which leaves me with a 2 I can't play on any 2.

	My best move is to play the 2, since it's the only one that will guarantee me the advantage in the next few rounds. I can force a reaction from the Landlord or draw the Red Joker if they have it, giving me the opportunity to win with the Joker after they play.

	So here is my plan: I play the 2. If the Landlord beats it, I can use the Joker to regain control. If they pass, then I lead the J next. It's a risk, but it\'s a win that will likely work since they're playing singles. The strategy to win is: Play small, beat small, then big to win. So, \ensuremath{\clubsuit}2 it is!

	\medskip
	Answer: Play \ensuremath{\clubsuit}2
\end{promptbox}

\begin{promptbox}{Fan Calculation: Mahjong (Gemini3-Pro-Preview)}
	\textbf{Case A: Win on B7:}[B4, B4, B4], [B6, B6, B6], [B7, B7, B7], [F2, F2, F2], [W6, W6]. Patterns: All Pungs (6 points), Seat Wind (2 points), and One Voided Suit (1 point) + Pung of Honors (1 point). Total: 10 points - this is a winner.

	\textbf{Case B: Win on W6:} [B4, B4, B4], [B6, B6, B6], [W6, W6, W6], [F2, F2, F2], [B7, B7]. Patterns: Same 10 points: All Pungs, Seat Wind, Voided Suit, and Honors.
\end{promptbox}

\section{Use of Large Language Model}
In the preparation of this manuscript, the author utilized an \ac{llm} for the purpose of text polishing and refinement. This includes improving grammar, sentence structure, and overall clarity. The authors remain fully responsible for the entire academic content, including all ideas, arguments, experiment methodologies, and conclusions presented herein.

\end{document}